\documentclass[review]{elsarticle}
\usepackage{geometry}
\newcommand{\tr}[1]{\textrm{#1}}
\newcommand{\mr}[1]{\mathrm{#1}}
\newcommand{\tnr}[1]{{\textnormal{#1}}}

\newcommand{\mc}[1]{\mathcal{#1}}
\newcommand{\mf}[1]{\mathsf{#1}}
 
\newcommand{\ms}[1]{\mathds{#1}}

\newcommand{\ov}[1]{\overline{#1}}

\newcommand{\bI}{\boldsymbol{I}}

\newcommand{\bx}{\boldsymbol{x}}

\newcommand{\bzero}{\boldsymbol{0}}
\newcommand{\bone}{\boldsymbol{1}}

\newcommand{\btheta}{\boldsymbol{\theta}}

\newcommand{\figref}[1]{Fig.~\ref{#1}}

\newcommand{\tabref}[1]{Table~\ref{#1}}

\newcommand{\ie}{i.e.,~} 		
\newcommand{\eg}{e.g.,~}	

\newcommand{\argmin}{\mathop{\mr{argmin}}}

\newcommand{\set}[1]{\{#1\}}
\newcommand{\SET}[1]{\left\{#1\right\}}

\newcommand{\ld}{\ldots}

\newcommand{\e}{\mr{e}}

\newcommand{\PR}[1]{\Pr\SET{#1}}       	

\newcommand{\Ex}{\ms{E}}     			
\newcommand{\T}{^{\mf{T}}}            		
\newcommand{\dd}{\,\mr{d}}             		

\newcommand{\mcL}{\mc{L}}

\newcommand{\mcN}{\mc{N}}

\newcommand{\matH}{\tnr{\textbf{H}}}
\newcommand{\matI}{\tnr{\textbf{I}}}

\newcommand{\matR}{\tnr{\textbf{R}}}

\usepackage[acronym,nonumberlist]{glossaries} 
\newacronym[\glsshortpluralkey=PDFs,\glslongpluralkey=probability density functions]{pdf}{PDF}{probability density function}
\newacronym[\glsshortpluralkey=CDFs,\glslongpluralkey=cumulative density functions]{cdf}{CDF}{cumulative density function}
\newacronym[\glsshortpluralkey=CCDFs,\glslongpluralkey=complementary cumulative density functions]{ccdf}{CDF}{complementary cumulative density function}
\newacronym[\glsshortpluralkey=PMFs,\glslongpluralkey=probability mass functions]{pmf}{PMF}{probability mass function}
\newacronym[]{lhs}{l.h.s.}{left-hand side}
\newacronym[]{rhs}{r.h.s.}{right-hand side} 

\newacronym[]{bicm}{BICM}{bit-interleaved coded modulation}
\newacronym[]{bicmid}{BICM-ID}{BICM with iterative demapping}
\newacronym[]{cm}{CM}{coded modulation}
\newacronym[]{tcm}{TCM}{trellis-coded modulation}
\newacronym[]{mlc}{MLC}{multi-level coding}
\newacronym[]{pam}{PAM}{pulse amplitude modulation}
\newacronym[]{bpsk}{BPSK}{binary phase shift keying}
\newacronym[]{qam}{QAM}{quadrature amplitude modulation}
\newacronym[]{16qam}{16-QAM}{16-points quadrature amplitude modulation}
\newacronym[]{psk}{PSK}{phase shift keying}
\newacronym[\glsshortpluralkey=LLRs,\glslongpluralkey=logarithmic likelihood ratios]{llr}{LLR}{logarithmic likelihood ratio}
\newacronym[]{oc}{OC}{operating characteristic}
\newacronym[]{dmp}{DMP}{discretized message passing}
\newacronym[]{mp}{MP}{message passing}
\newacronym[]{ep}{EP}{expectation propagation}
\newacronym[\glsshortpluralkey=MIs,\glslongpluralkey=mutual informations]{mi}{MI}{mutual information}
\newacronym[\glsshortpluralkey=GMIs,\glslongpluralkey=generalized mutual informations]{gmi}{GMI}{generalized mutual information}
\newacronym[]{eesm}{EESM}{exponential effective-SNR-mapping}
\newacronym[]{bicm-gmi}{BICM-GMI}{BICM generalized mutual information}
\newacronym[]{awgn}{AWGN}{additive white Gaussian noise}
\newacronym[]{bsc}{BSC}{binary symetric channel}
\newacronym[]{amc}{AMC}{adaptive modulation and coding}
\newacronym[]{csi}{CSI}{channel state information}
\newacronym[]{cqi}{CQI}{channel quality indicator}
\newacronym[]{kl}{KL}{Kullback-Leibler}
\newacronym[]{cmm}{CMM}{circular moment matching}
\newacronym[]{ga}{GA}{Gaussian approximation}

\newacronym[]{sp}{SP}{set-partitioning}
\newacronym[]{gsm}{GSM}{global system for mobile communications}
\newacronym[]{edge}{EDGE}{enhanced data rates for GSM evolution}
\newacronym[]{3gpp}{3GPP}{3rd generation partnership project}
\newacronym[]{umts}{UMTS}{Universal Mobile Telecommunication System}
\newacronym[]{lte}{LTE}{Long Term Evolution}
\newacronym[]{dvb}{DVB}{digital video broadcasting}
\newacronym[]{fdd}{FDD}{Frequency Division Duplexing}

\newacronym[\glsshortpluralkey=CCs,\glslongpluralkey=convolutional codes]{cc}{CC}{convolutional code}
\newacronym[\glsshortpluralkey=PCCCs,\glslongpluralkey=parallel concatenated convolutional codes]{pccc}{PCCC}{parallel concatenated convolutional code}
\newacronym[\glsshortpluralkey=TCs,\glslongpluralkey=turbo codes]{tc}{TC}{turbo code}
\newacronym{ldpc}{LDPC}{low-density parity-check}
\newacronym[]{ofdm}{OFDM}{orthogonal frequency-division multiplexing}
\newacronym[]{bep}{BEP}{bit-error probability}
\newacronym[]{wep}{WEP}{word-error probability}
\newacronym[]{sep}{SEP}{symbol-error probability}
\newacronym[]{pep}{PEP}{pairwise-error probability}
\newacronym[]{ttcm}{TTCM}{turbo-trellis coded modulation}
\newacronym[]{uep}{UEP}{unequal error protection}
\newacronym[\glsshortpluralkey=CENCs,\glslongpluralkey=convolutional encoders]{cenc}{CENC}{convolutional encoder}
\newacronym[]{mimo}{MIMO}{multiple-input multiple-output}
\newacronym[\glsshortpluralkey=SNRs,\glslongpluralkey=signal-to-noise ratios]{snr}{SNR}{signal-to-noise ratio}
\newacronym[\glsshortpluralkey=SINRs,\glslongpluralkey=the signal-to-interference-plus-noise ratios]{sinr}{SINR}{the signal-to-interference-plus-noise ratio}
\newacronym[]{msb}{MSB}{most-significative bit}
\newacronym[]{bcjr}{BCJR}{Bahl--Cocke--Jelinek--Raviv}
\newacronym[]{cbc}{CBC}{Colavolpe--Barbieri--Caire}
\newacronym[]{skr}{SKR}{Shayovitz--Kreimer--Raphaeli}
\newacronym[\glsshortpluralkey=SEDs,\glslongpluralkey=squared Euclidean distances]{sed}{SED}{squared Euclidean distance}
\newacronym[\glsshortpluralkey=EDs,\glslongpluralkey=Euclidean distances]{ed}{ED}{Euclidean distance}
\newacronym[\glsshortpluralkey=MEDs,\glslongpluralkey=minimum Euclidean distances]{med}{MED}{minimum Euclidean distance}
\newacronym[]{core}{CoRe}{constellation rearrangement}
\newacronym[]{pdl}{PDL}{parallel decoding of the individual levels}
\newacronym[\glsshortpluralkey=GCs,\glslongpluralkey=Gray codes]{gc}{GC}{Gray code}
\newacronym[]{brgc}{BRGC}{binary-reflected Gray code}
\newacronym[]{nbc}{NBC}{natural binary code}
\newacronym[]{fbc}{FBC}{folded-binary code}
\newacronym[]{bsgc}{BSGC}{binary semi-Gray code}
\newacronym[]{msp}{MSP}{modified set-partitioning}
\newacronym[]{ssp}{SSP}{semi set-partitioning}
\newacronym[]{fhd}{FHD}{free Hamming distance}
\newacronym[]{mfhd}{MFHD}{maximum free Hamming distance}
\newacronym[]{ods}{ODS}{optimal distance spectrum}
\newacronym[]{iud}{i.u.d.}{independent and uniformly distributed}
\newacronym[]{ud}{u.d.}{uniformly distributed}
\newacronym[]{iid}{i.i.d.}{independent, identically distributed}
\newacronym[]{ami}{AMI}{accumulated mutual information}
\newacronym[]{bico}{BICO}{binary-input continuous-output}
\newacronym[]{gh}{GH}{Gauss--Hermite}
\newacronym[]{gum}{GUM}{Gaussian--uniform mixture}

\newacronym[\glsshortpluralkey=BSs,\glslongpluralkey=base-stations]{bs}{BS}{base-station}
\newacronym[\glsshortpluralkey=MSs,\glslongpluralkey=mobile-stations]{ms}{MS}{mobile-stations}

\newacronym[]{phy}{PHY}{physical layer} 
\newacronym[]{rlc}{RLC}{Radio-Link control} 
\newacronym[]{ran}{RAN}{Radio Access Network} 
\newacronym[]{llc}{LLC}{logical link control} 
\newacronym[]{tcp}{TCP}{transmission control protocol} 
\newacronym[]{mac}{MAC}{media access control} 
\newacronym[]{fft}{FFT}{fast Fourier transform} 
\newacronym[]{ft}{FT}{Fourrier transform}
\newacronym[]{cf}{CF}{characteristic function} 
\newacronym[]{mgf}{MGF}{moment generating function} 
\newacronym[]{ee}{EE}{energy efficiency} 
\newacronym[]{eb}{EB}{energy per bit}
\newacronym[]{kkt}{KKT}{Karush--Kuhn--Tucker} 
\newacronym[]{mcs}{MCS}{modulation/coding scheme} 
\newacronym[]{fec}{FEC}{forward error correction}
\newacronym[]{arq}{ARQ}{automatic repeat request}
\newacronym[]{harq}{HARQ}{hybrid ARQ}
\newacronym[]{tarq}{TARQ}{truncated HARQ}
\newacronym[]{ir}{IR}{incremental redundancy}
\newacronym[]{rpr}{RR}{repetition redundancy}
\newacronym[]{rrharq}{RR-HARQ}{repetition redundancy HARQ}
\newacronym[]{irharq}{IR-HARQ}{incremental redundancy HARQ}
\newacronym[]{ack}{ACK}{positive acknowledgment}
\newacronym[]{nack}{NACK}{negative acknowledgment}
\newacronym[]{hol}{HoL}{head of the line}
\newacronym[]{crc}{CRC}{cyclic redundancy check}
\newacronym[]{dp}{DP}{dynamic programming}
\newacronym[]{gp}{GP}{geometric programming}
\newacronym[]{per}{PER}{packet error rate}
\newacronym[]{ber}{BER}{bit error rate}
\newacronym[]{op}{OP}{outage probability}
\newacronym[]{spa}{SPA}{saddle-point approximation}
\newacronym[]{mrc}{MRC}{maximum ratio combining}
\newacronym[]{mdp}{MDP}{Markov decision process}
\newacronym[]{lp}{LP}{linear programming}
\newacronym[]{pomdp}{POMDP}{partially observable Markov decision process}
\newacronym[]{psimdp}{PSI-MDP}{partial state information Markov decision process}
\newacronym[]{scpp}{SCPP}{stochastic shortest path problem}

\newacronym[]{forw}{frwd}{forward}
\newacronym[]{feed}{fdbk}{feedback}

\newacronym[]{mm}{MM-HARQ}{multi-message HARQ}
\newacronym[]{xp}{XP-HARQ}{cross-packet HARQ}
\newacronym[]{ts}{TS}{time-sharing}
\newacronym[]{sc}{SC}{superposition coding}
\newacronym[]{sbrq}{SBRQ}{systematic backward retransmission}
\newacronym[]{brq}{BRQ}{backward retransmission}
\newacronym[]{lharq}{L-HARQ}{layer-coded HARQ}
\newacronym[]{anlharq}{AoN-HARQ}{all-or-none L-HARQ}
\newacronym[]{vlharq}{VL-HARQ}{variable-length HARQ}

\newacronym[]{pp}{PPP}{point process}
\newacronym[]{ppp}{PPP}{Poisson point process}

\newacronym[]{fide}{FIDE}{F\'ed\'eration Internationale des \'Echecs}
\newacronym[]{fifa}{FIFA}{F\'ed\'eration Internationale de Football Association}
\newacronym[]{fivb}{FIVB}{F\'ed\'eration Internationale de Volleyball}
\newacronym[]{epl}{EPL}{English Premier League}
\newacronym[]{nhl}{NHL}{National Hockey League}
\newacronym[]{nfl}{NFL}{National Football League}

\newacronym[]{sg}{SG}{stochastic gradient}
\newacronym[]{lms}{LMS}{least mean squares}
\newacronym[]{rls}{RLS}{recursive least squares}
\newacronym[]{vss}{VSS}{variable step-size}
\newacronym[]{hfa}{HFA}{home-field advantage}
\newacronym[]{ha}{HA}{home advantage}
\newacronym[]{mov}{MOV}{margin of victory}
\newacronym[]{ac}{AC}{Adjacent Categories}
\newacronym[]{cl}{CL}{Cumulative Link}
\newacronym[]{rps}{RPS}{Ranked Probability Score}
\newacronym[]{mse}{MSE}{Mean Squared Error}
\newacronym[]{mmse}{MMSE}{Minimum Mean Squared Error}
\newacronym[]{rmse}{RMSE}{Root Mean Squared Error}
\newacronym[]{map}{MAP}{maximum a posteriori}
\newacronym[]{ml}{ML}{maximum likelihood}
\newacronym[]{loo}{LOO}{leave-one-out}
\newacronym[]{alo}{ALO}{approximate leave-one-out}
\newacronym[]{msd}{MSD}{mean-square deviation}

\newacronym[]{svd}{SVD}{singular values decomposition}

\newacronym[]{skf}{SKF}{Simplified Kalman Filter}
\newacronym[]{vskf}{vSKF}{\emph{vector-covariance} Simplified Kalman Filter}
\newacronym[]{sskf}{sSKF}{\emph{scalar-covariance} Simplified Kalman Filter}
\newacronym[]{fskf}{fSKF}{\emph{fixed-variance} Simplified Kalman Filter}
\newacronym[]{kf}{KF}{Kalman Filter}
\newacronym[]{gelo}{G-Elo}{Generalized Elo}

\newacronym[]{tpb}{TPB}{tensor-product-basis}

\usepackage{amsmath,amssymb,amscd,latexsym,dsfont,wasysym,bbm}
\usepackage{natbib}  
\biboptions{numbers,sort&compress}
\usepackage{amsmath} 
\usepackage[caption=false,font=footnotesize]{subfig} 
\usepackage[normalem]{ulem} 

\usepackage{enumitem}
\newlist{Aenumerate}{enumerate}{1}
\setlist[Aenumerate]{label=A.\arabic*}

\newcommand\numberthis{\addtocounter{equation}{1}\tag{\theequation}}

\usepackage{booktabs}

\usepackage{microtype}

\usepackage{hyperref}

\usepackage{empheq}

\usepackage{xcolor}
\definecolor{cblue}{HTML}{1965B0}

\definecolor{cred}{HTML}{B8221E}

\definecolor{dgreen}{rgb}{0,0.6,0}

\definecolor{dorange}{RGB}{255, 128, 0}

\definecolor{burntorange}{rgb}{0.8, 0.33, 0.0}

\begin{document}

\title{Stochastic analysis of the Elo rating algorithm\\ in round-robin tournaments}

\author[1,2]{Daniel~Gomes~de~Pinho~Zanco}
\ead{daniel.zanco@inrs.ca}

\author[2]{Leszek~Szczecinski}
\ead{leszek@emt.inrs.ca}

\author[1,3]{Eduardo~Vinicius~Kuhn\corref{cor1}}
\ead{kuhn@utfpr.edu.br}

\author[1]{Rui~Seara}
\ead{seara@linse.ufsc.br}

\cortext[cor1]{Corresponding author.}
\address[1]{LINSE-Circuits and Signal Processing Laboratory, Department of Electrical and Electronics Engineering, Federal University of Santa Catarina, Florian\'opolis, Santa Catarina, 88040-900, Brazil.}

\address[2]{INRS–Institut National de la Recherche  Scientific, Montreal, QC, H5A-1K6, Canada.}

\address[3]{LAPSE–Electronics and Signal Processing Laboratory, Department of Electronics Engineering, Federal University of Technology - Paran\'a, Toledo, Paran\'a, 85902-490, Brazil.}

\begin{abstract}
The Elo algorithm, renowned for its simplicity, is widely used for rating in sports tournaments and other applications. However, despite its widespread use, a detailed understanding of the convergence characteristics of the Elo algorithm is still lacking. Aiming to fill this gap, this paper presents a comprehensive (stochastic) analysis of the Elo algorithm, considering round-robin tournaments. Specifically, analytical expressions are derived describing the evolution of the skills and performance metrics. Then, taking into account the relationship between the behavior of the algorithm and the step-size value, which is a hyperparameter that can be controlled, design guidelines and discussions about the performance of the algorithm are provided. Experimental results are shown confirming the accuracy of the analysis and illustrating the applicability of the theoretical findings using real-world data obtained from SuperLega, the Italian volleyball league.
\end{abstract}


\journal{Digital Signal Processing.}

\let\today\relax

\maketitle
\section{Introduction}

Over the recent decades, different ranking/rating algorithms have been devised and used in tournaments/competitions (sports and games) to assign numbers to teams (or players in the case of individual sports) \cite{stefani2011}. The rating consists of assigning a numerical value to a competitor based on empirical observations (of the game outcomes), while the ranking is obtained by sorting the rating values \cite{stefani2011}; so, a ``stronger" team has a large rating and is ranked before the weaker team \cite{barrow2013}. Hence, the rating, which reflects the skills/abilities/strength of each team compared to the others (see the extensive surveys in \cite{stefani2011,barrow2013,lasek2021,Csato2021}), can be used to obtain quick information on the state of a tournament, to assess the strength of teams, to decide on seeding/pairing (defining game schedules with more ``interesting'' matches) or promotion/relegation (between high- and low-level leagues) \cite{lasek2021}, as well as to provide support for both bookmakers and gamblers \cite{hvattum2010,wolf2020}. As a consequence of practical applicability, the development and refinement of rating\footnote{The problem of rating falls into the topic of ``paired comparison'' modeling in the field of statistics \cite{BradleyTerry52,Glickman93_thesis,leitner2010}.} systems have become an important and active research field (as properly summarized, \eg through \cite{vaughanwilliams2010} and \cite{mchale2019} and references therein).

From a statistical perspective, the rating consists of inferring unknown parameters (skills) from the observed outcomes (game results) via a formal mathematical model of skills-outcomes (see \cite[Ch.~4]{dobson2011} for details on goal- or result-based models, and \cite[Appendix]{stekler2010} for a sport-by-sport discussion). In this sense, the Elo algorithm (or Elo rating system)\footnote{Named in honor of its originator, Arpad Emmerich Elo, who was a Hungarian-American physicist and professor \cite{Elo78_Book}.}, devised in the 1960s to rate chess players \cite{Elo78_Book}, is arguably the most well-known and widely used, mainly due to its simplicity and intuitive interpretation \cite{lasek2021}. This algorithm assumes that each team has a ``single'' parameter governing its performance (\ie ``its true strength'') and then the goal is to provide an estimate of such a parameter (\ie ``a rating"); to this end, the algorithm increases the skills of a particular team when the score of the game is larger than the expected/predicted one and vice versa \cite{Elo78_Book}, thus updating (after each match) the rating of the teams depending on the game outcomes \cite{barrow2013,ryall2010}. Therefore, the self-correcting characteristic of the algorithm allows it to (intrinsically) track the skills of the teams over time \cite{glickman1995}.

The Elo algorithm, originally used for rating in chess by \gls{fide} \cite{fide}, has become a de facto standard for rating and has been adapted, most notably, by \gls{fifa} for ranking of the association football\footnote{For an overview of the most common ranking methods, see \cite{Eetvelde2019}.} (soccer) national Women's teams in 2003 \cite{fifa_rating_W} and Men's teams in 2018 \cite{fifa_ranking}. Variants of this algorithm have also been applied to other sports \cite{play-pokemon,Silver14,carbone2016,FiveThirtyEight,Glickman1999,Herbrich06,leitner2010,ryall2010,lasek2013,dorsey2019,kovalchik2020,wolf2020,Szczecinski20,lasek2021,wheatcroft2021,Szczecinski22,Szczecinski_FIFA2022,eloratings.net}, as well as in the context of tournament simulations \cite{Chater2021,Csato2022,Csato2023}. Although the rating principles are very similar, each application can use its own version of the algorithm, \eg incorporating custom step sizes, the so-called \gls{hfa}, and/or different approaches for modeling the game outcomes. Especially regarding the choice of the step-size value (which is the only hyperparameter of the Elo algorithm), it is usually made in a heuristic manner and differs significantly from one application to another (\eg see \cite[Table~10]{stefani2011} and \cite[Sec.~3]{lasek2013}).

Despite the widespread use of the Elo algorithm, the discussions presented so far in the literature have mostly been based on empirical studies. In fact, to the best of our knowledge, a theoretical analysis of convergence has only been discussed in \cite{jabin2015}, in which a continuous and deterministic kinetic model based on the mean-field theory has been used to study the evolution of the ratings in a round-robin tournament. In this context, the authors have demonstrated, through \cite[Th.~2.2]{jabin2015}, that the estimates of skills converge to their ``true strengths" at an exponential rate in tournaments with a large number of players. Apart from this, neither practical guidelines for hyperparameter tuning nor empirical validation using real-world data are presented. Therefore, as already pointed out critically by \cite{aldous2017,kovalchik2020}, further theoretical studies of the Elo algorithm and its variants would still be valuable.

In this context, the present research work aims to obtain a stochastic model for characterizing the Elo algorithm, considering round-robin tournaments. In particular, aiming to gain more insights into the algorithm and avoid resorting to extensive numerical simulations, our approach relies on mathematical tools adapted from the area of adaptive filters \cite{sayed2008,farhang2013,haykin2014,tobias2004,kuhn2014,matsuo2019,matsuo2021,bakri2022,bakri2022b}. Specifically, we aim to
\begin{enumerate}[label=\roman*)]
\item provide a comprehensive mathematical framework to analyze the Elo algorithm applied in a round-robin tournament;

\item obtain analytical expressions characterizing the evolution of skills (along with other important performance metrics) throughout a tournament;

\item investigate the impact of the hyperparameters on the performance of the algorithm, aiming to obtain useful insights and design guidelines; and

\item compare the analytical results with those obtained through the algorithm using real-world data (from the Italian SuperLega, Volleyball League), illustrating the applicability of the theoretical findings.
\end{enumerate}

The remainder of this paper is organized as follows. Section~\ref{Sec:Model} revisits some fundamental concepts related to the rating model and the Elo algorithm. Section~\ref{Sec:analysis} presents the proposed stochastic analysis for the Elo algorithm while Section~\ref{Sec:Discussion} discusses some interesting insights and design guidelines obtained from the analysis. Section~\ref{Sec:experimental_results} presents experimental results based on data extracted from the Italian league of volleyball. Lastly, conclusions and suggestions for future research works are summarized in Section~\ref{Sec:Conclusions}.

The mathematical notation adopted in this paper follows the standard practice of using lower-case boldface letters for vectors, upper-case boldface letters for matrices, and italic letters for scalar quantities. All involved variables are assumed real-valued. Superscript $\mf{T}$ stands for the transpose operator, $\Ex(\,\cdot\,)$ denotes the expected value operator which (unless otherwise stated explicitly) is calculated with respect to all random variables appearing inside parenthesis/brackets, while $[\mathbf{A}]_{m,n}$ represents the scalar $m,n$-entry of matrix $\mathbf{A}$. Still, when dealing with probabilities, we abuse the notation to write simply $\PR{a|b}$ instead of $\PR{A=a|b}$ with $A$ denoting the random variable itself and $a$ its realization conditioned on $b$. Table~\ref{tab:symbols} provides a brief summary of the variables and definitions that are central to the discussion in this paper, while secondary variables of contextual relevance (not listed here) are defined throughout the text.

\begin{table}[ht]
\centering
\caption{Summary of variables and their definitions used in this paper.}
\label{tab:symbols}
\medskip
\begin{tabular}{@{}c l@{}}
\toprule
Variable & Definition  \\ \midrule
$M$             & Number of teams/players \\
$K$             & Total number of games/matches \\
$k$             & Match/game index \\
$\btheta$       & Vector containing the skills of the teams/players\\
$\btheta^*$     & Vector containing the hypothetical ``true'' skills \\
$\btheta_k$     & Vector containing the estimate of the skills\\
$\bx_k$         & Scheduling vector\\
$z_k$           & Difference between the skills of the opponents\\
$y_k$ & Match/game outcome\\
$\beta$         & Step-size parameter \\
$\eta$          & Home-field advantage (HFA)\\
$v$             & Variance of $\btheta^*$ \\
$\sigma(\cdot)$ & Logistic function \\
$\mcL(\cdot)$   & Probability density function of the logistic distribution\\
$\ell_k(\btheta)$ & Negative log-likelihood function (loss function) \\
$\matR$         & Autocorrelation matrix of $\bx_k$ \\
$\ov{d}_k$      & Mean-square deviation (MSD) of the skills \\
$\ov\ell_k$     & Mean behavior of the loss function \\
$\ov\ell_{\rm min}$ & Lower bound on the mean of the loss function \\
$\ov\ell_{{\rm ex},k}$ & Mean excess-loss \\
$\alpha_1$      & Constant of convergence in the mean sense \\
$\alpha_2$      & Constant of convergence in the mean-square sense \\
$\tau_1$        & Convergence time constant in the mean sense \\
$\tau_2$        & Convergence time constant in the mean-square sense \\
$v_\tr{th}$     & A boundary that separates small and large $v$ \\
$\beta_{\tr{o}, k}$ & Optimum step-size value for $k$ games/matches\\
\bottomrule
\end{tabular}
\end{table}

\section{Problem statement}\label{Sec:Model}

Here, we first introduce the rating model along with some fundamental concepts, and then present a recursive algorithm that is used to solve the rating problem.

\subsection{Rating model}
Let us consider $M$ teams (with index $m=1,\ld, M$) playing against each other in a round-robin tournament/competition (see \citep{Szymanski2003,scarf2006,Scarf2009,Lasek2018,Csato2021} for detailed discussions on tournament design) with games indexed through $k=1,\ld,K$, where $K$ is the total number of games observed in a given period of time (\eg the number of games in a season). In each game $k$, opponents are scheduled by an external mechanism that assigns ``home'' and ``away'' indices ($i_k$ and $j_k$, respectively) to the teams. The distinction between home and away teams is particularly useful in games in which \gls{hfa}\footnote{It is worth mentioning that the advantage is not always tied to the physical field of the game; in fact, the concept of the home-field advantage (HFA) is domain-specific and can be influenced by other factors (e.g., the advantage conferred by the white pieces in chess \cite{GonzalezDiaz2016}).} is present, \ie when playing at home increases the probability of winning (for details, see \cite[Sec.~2]{ryall2010}). The results of the games are denoted as $y_k \in \{0,1\}$ with $y_k = 1$ when the home team wins and $y_k = 0$ when the away team wins. Note that the extension to multi-level results (\eg including draws) requires different approaches for modeling\footnote{For instance, the formal probabilistic model of draws leads to the so-called Elo--Davidson algorithm, which, under a particular choice of parameters, is identical to the Elo algorithm \cite{Szczecinski20}.} the game outcomes, leading to modifications of the algorithm itself.

In the rating problem, we are interested in finding the skills of the teams $\btheta = [\theta_1,\ld,\theta_M]\T$ that make it possible to directly compare the abilities of the $M$ teams. Statistical inference of skills $\btheta$ from observed outcomes $y_k$ requires a model that relates both quantities. To this end, we adopt the popular Bradley--Terry model \cite{BradleyTerry52}
\begin{align}\label{Pr.y.1}
\PR{y_k = 1|\btheta} &= \sigma(z_k/s + \eta)
\end{align}
which relies on the difference between the skills of the playing teams
\begin{align*}
z_k &= \theta_{i_k} - \theta_{j_k}\\
    &= \bx_k\T \btheta \numberthis \label{z_k}
\end{align*}
with the scheduling vector $\bx_k=[x_{k,1}\;\ldots\; x_{k,M}]\T$ defined as
\begin{align} \label{eq:schedule_vector}
x_{k,m} =
\begin{cases}
1,&  m = i_k \\
-1,& m = j_k \\
0,&  \text{otherwise}
\end{cases}
\end{align}
while
\begin{align}\label{logit}
\sigma(z) = \frac{1}{1+\e^{-z}}
\end{align}
is the logistic function and $\eta$ plays the role of the \gls{hfa} parameter (capturing the increased probability of the home-team winning).

Note that $s > 0$ in \eqref{Pr.y.1} and the exponent base in \eqref{logit} are arbitrary scaling factors used to stretch/accommodate the value of the skills to a desirable range, \eg
\begin{align*}
\sigma(z/s) &= \frac{1}{1+\e^{-z/s}}\\
&= \frac{1}{1+10^{-z/s'}}\numberthis \label{logit10}
\end{align*}
where
\begin{align}
\label{scale-ln10}
s' &= s\ln(10).
\end{align}
Therefore, \eqref{logit10} links the logistic function \eqref{logit} to the base-10 function commonly used in the Elo rating, ratifying that the change of scale can be seen as a change of the base of the exponent (as also noted in \cite{leitner2010}); henceforth, for the sake of simplicity, we consider $s = 1$.

\subsection{Rating problem and algorithm}\label{Sec:rating.problem}

Taking into account the total number of games $K$, as well as assuming that the home-field advantage $\eta$ and the scheduling vectors $\bx_k$ are known, the problem at hand is to estimate the skills $\btheta$ from the game outcomes $y_k$. To solve this problem, which can be interpreted as a logistic regression problem \cite[Ch.~4.4]{Hastie_book}, we can resort to the \gls{ml} principle to write
\begin{align}\label{ML.optimization}
\hat\btheta = \argmin_{\btheta}\; \sum_{k=1}^{K} \ell_k(\btheta)
\end{align}
with the negative log-likelihood function (also known as loss function) given by
\begin{align}\label{eq:loss_k}
\nonumber
\ell_k(\btheta) &= -\ln[\PR{y_k | \btheta}]\\
&= -y_k \ln[\sigma(\bx_k\T\btheta+\eta)] - (1 - y_k) \ln[1 - \sigma(\bx_k\T\btheta+\eta)].
\end{align}
A solution for \eqref{ML.optimization} can be obtained recursively through the \gls{sg} optimization method (as described in \cite{lasek2021}), \ie
\begin{align}
\nonumber
\btheta_{k+1} &= \btheta_{k} - \beta \nabla \ell_k(\btheta_k) \\
\label{eq:rating-sgd2}
&= \btheta_{k} + \beta [y_k - \sigma(\bx_k\T \btheta_k+\eta)] \bx_k
\end{align}
with $\beta > 0$ denoting the step-size parameter and $\btheta_0$ being conveniently initialized as an $M$-dimensional vector of zeros. In practice, $\beta$ and $\btheta_0$ may depend on the particulars of each application, \eg \gls{fifa} ranking adjusts $\beta$ depending on the type of the game (small in friendlies but large in the World Cup) and $\btheta_0$ based on the existing ranking positions on its initial deployment \cite{Szczecinski_FIFA2022}. Note that \eqref{eq:rating-sgd2} represents exactly the Elo algorithm \cite{Elo78_Book} and describes how the ratings are updated after each game $k$ (for practical and illustrative examples, see \cite[Ch.~5]{Langeville12_book}).

\section{Proposed stochastic analysis}\label{Sec:analysis}
In this section, a stochastic analysis of the Elo rating algorithm is presented. Specifically, we first introduce some statistical assumptions along with an approximate version of the rating algorithm, which are required to make the mathematical development tractable; next, mathematical expressions are derived to characterize the mean behavior of the skills, the mean-square deviation of the skills, the variance of the skills, and the mean behavior of the loss function.

\subsection{Assumptions}\label{Sec:assumptions}

Since the rating algorithm [given by \eqref{eq:rating-sgd2}] depends on the realizations of random variables, its analysis (in the stochastic sense) becomes feasible under the following assumptions:

First, although the sequence of scheduling vectors $\bx_k$ [see \eqref{eq:schedule_vector}] in round-robin tournaments is determined in advance, we assume that the indices $i_k$ and $j_k$ (which define $\bx_k$) are random, drawn uniformly and without repetition from the set $\set{1,\ld,M}$ (\ie each team plays at home and away against all other teams with the same probability) so that
\begin{align} \label{eq:PR.schedules}
\nonumber
\PR{i_k = m} &= \PR{j_k = m} ,\quad \forall m\\
&= \frac{1}{M}
\end{align}
and
\begin{align} \label{eq:PR.schedules.cond}
\nonumber
\PR{j_k = n | i_k = m} &= \PR{i_k = n | j_k = m}, \quad \forall m \neq n\\
&= \frac{1}{M - 1}.
\end{align}
Thus, the correlation between the elements of $\bx_k$ can be expressed as
\begin{align} \label{eq:R.elements.generic}
\Ex[x_{k,m} x_{k,n}]
&=
\begin{cases}
\PR{i_k = m} + \PR{j_k = m}, &m = n\\
-\PR{i_k = m} \PR{j_k = n | i_k = m}  -\PR{j_k = m} \PR{i_k = n | j_k = m}, &\text{otherwise}
\end{cases}
\end{align}
which results [from \eqref{eq:PR.schedules} and \eqref{eq:PR.schedules.cond}] in
\begin{align} \label{eq:R.elements.uniform}
\Ex[x_{k,m} x_{k,n}]
&=
\begin{cases}
\displaystyle\frac{2}{M}, & m = n \\
-\displaystyle\frac{2}{M (M-1)},\quad  &\text{otherwise.}
\end{cases}
\end{align}
So, the autocorrelation matrix of $\bx_k$ can be compactly represented as
\begin{align}
\label{covariance.matrix}
\nonumber
\matR &= \Ex[\bx_k\bx_k\T]\\
      &= \frac{2}{M-1} \left(\bI - \frac{1}{M}\bone\bone\T\right)
\end{align}
which holds as $M \rightarrow \infty$.

Second, we assume that the Bradley--Terry model \eqref{Pr.y.1} describes exactly the distribution of $y_k$ when conditioned on hypothetical ``true'' skills $\btheta^*= [\theta_1^*,\ld,\theta_M^*]\T$, \ie
\begin{align} \label{Pr.y.star}
\PR{y_k=1|\btheta^*} &= \sigma(z_k^* + \eta)
\end{align}
in which
\begin{align*}
z_k^* &= \theta_{i_k}^* - \theta_{j_k}^*\\
&= \bx_k\T \btheta^* \numberthis \label{z_k*}
\end{align*}
with the scheduling vector $\bx_k$ given by \eqref{eq:schedule_vector}. Note in \eqref{Pr.y.star} that we have already assumed $s=1$ for simplicity.

Third, to remove the dependence on unknown $\btheta^*$, we assume that its elements are drawn from a Gaussian distribution with zero mean and variance $v$ as in \cite[Sec.~4.1]{Bock1972} and \cite{Gramacy13}. As a consequence of this assumption on $\btheta^*$, the characteristics of a given tournament are uniquely defined by $v$ whose value refers to different contexts of competition. In particular, for $\eta=0$, large $v$ implies that the difference in skills between opponents $z_k^*$ [see \eqref{z_k*}] is likely to be large and, then, the game result $y_k$ [generated randomly according to \eqref{Pr.y.star}] will be quasi-deterministic, \ie either $\PR{y_k=1|\btheta^*} \approx 1$ or $\PR{y_k=1|\btheta^*} \approx 0$. On the other hand, a small value of $v$ leads to $z_k^* \approx 0$ which means that the uncertainty of the results conditioned on skills is high, \ie $\PR{y_k=1|\btheta^*} \approx 0.5$. As practical examples, consider that i) in a professional league, where teams are expected to perform similarly, a significant uncertainty in game results is then modeled by small $v$; while ii) for international tournaments/championships, where large disparities in player quality are observed, games between the strongest and weakest teams are usually quite easy to predict, which implies large $v$. 

Note that the assumptions introduced here regarding the scheduling vector $\bx_k$ and the vector of true skills $\btheta^*$ are somehow strong and debatable; nevertheless, such assumptions are required to make mathematical development tractable, leading to satisfactory results (as shown in Section~\ref{Sec:experimental_results}).

\subsection{Quadratic approximation of the rating problem}\label{Sec:approx.likelihood}

Analyzing \eqref{eq:rating-sgd2} is challenging due to the impossibility of isolating $\btheta_k$ on the \gls{rhs}, since the presence of the term $\sigma(\bx\T\btheta_k+\eta)$ makes the equation transcendental. So, aiming to make the mathematical development tractable, let us approximate \eqref{eq:loss_k} using a second-order Taylor series expansion around $\btheta^*$ as
\begin{align*}
\tilde{\ell}_k(\btheta) &= \ell_k(\btheta^*) + [\nabla \ell_k(\btheta^*)]\T (\btheta - \btheta^*) + \frac{1}{2} (\btheta - \btheta^*)\T \nabla^2 \ell_k(\btheta^*) (\btheta - \btheta^*)\\
\numberthis \label{eq:cost-taylor}
&= \ell_k(\btheta^*) + g_k\bx\T_k (\btheta - \btheta^*) + \frac{1}{2} h_k(\btheta - \btheta^*)\T \bx_k\bx_k\T (\btheta - \btheta^*)
\end{align*}
with
\begin{align}
\label{eq:gradient-optimal}
g_k &= y_k - \sigma(\bx_k\T\btheta^*+\eta)
\end{align}
and
\begin{align}
\label{eq:hessian-optimal2}
h_k &= \mcL(\bx_k\T \btheta^*+\eta)
\end{align}
where
\begin{align*}
\mcL(z) &= \sigma(z) [1 - \sigma(z)]\\
\numberthis \label{logistic.pdf}
&= \sigma(z) \sigma(-z)
\end{align*}
is the \gls{pdf} of the logistic distribution. Thereby, \eqref{eq:rating-sgd2} can be simplified by using \eqref{eq:cost-taylor} as
\begin{align*}
\btheta_{k+1} 
&\approx \btheta_{k}-\beta\nabla\tilde\ell(\btheta_k)\\
\numberthis \label{eq:sgd-taylor.2}
&=\btheta_{k} - \beta h_k \bx_k\bx_k\T (\btheta_k - \btheta^*) + \beta g_k \bx_k.
\end{align*}

\subsection{Mean behavior of the skills}\label{Sec:skills.mean}

To characterize the mean behavior of the skills, we start by taking the expected value of both sides of \eqref{eq:sgd-taylor.2} such that
\begin{align} \label{eq:mean-weight}
\Ex[\btheta_{k+1}] &= \Ex[\btheta_k] - \beta \Ex[h_k \bx_k\bx_k\T \btheta_k] + \beta \Ex[h_k \bx_k\bx_k\T] \btheta^* + \beta \Ex[g_k \bx_k].
\end{align}
Then, recalling that $\bx_k$ is statistically independent from $\btheta_k$, the second and third-terms of the \gls{rhs} of \eqref{eq:mean-weight} can be rewritten as
\begin{align*}
\Ex[h_k \bx_k\bx_k\T \btheta_k] &= \Ex[h_k \bx_k\bx_k\T]\Ex[\btheta_k]\\
\numberthis \label{eq:independence-assumption}
&= \matH \Ex[\btheta_k]
\end{align*}
and
\begin{align} \label{eq:H_star}
\Ex[h_k \bx_k\bx_k\T] \btheta^* = \matH\btheta^*
\end{align}
where
\begin{align}\label{matrix_H}
\matH = \tr{diag}(\matH' \bone) - \matH'
\end{align}
is the Fisher information matrix \cite[Sec.~3.7]{book:mkay} whose elements are given by
\begin{align}
[\matH']_{m,n} &= \frac{1}{M(M-1)} [ \mcL(\theta^*_{m}-\theta^*_{n}+\eta)+\mcL(\theta^*_{n}-\theta^*_{m}+\eta) ].
\end{align}
In turn, due to the orthogonality principle \cite[Sec.~3.3]{farhang2013}, the last term of the \gls{rhs} of \eqref{eq:mean-weight} reduces to
\begin{align*}
\Ex[g_k \bx_k] 
&= \Ex_{\bx_k,y_k}\{ [y_k - \sigma(\bx_k\T\btheta^* + \eta)] \bx_k \}\\
\numberthis \label{eq:g_kx_k}
&= \Ex_{\bx_k}\{ [\sigma(\bx_k\T \btheta^* + \eta)  - \sigma(\bx_k\T \btheta^* + \eta)] \bx_k \}\\ 
&= \bzero.
\end{align*}
Finally, substituting \eqref{eq:independence-assumption}, \eqref{eq:H_star}, and \eqref{eq:g_kx_k} into \eqref{eq:mean-weight}, we have
\begin{align}
\label{eq:weight-behavior}
\Ex[\btheta_{k+1}] = \Ex[\btheta_k] - \beta \matH (\Ex[\btheta_k] - \btheta^*).
\end{align}

Since \eqref{eq:weight-behavior} depends on the complete knowledge of $\btheta^*$ to determine $\matH$, we resort to the ergodicity argument to approximate the sample statistics of $\btheta^*$ [in \eqref{matrix_H}] to its expected value (which holds for a large number of players $M$). Thereby, $\matH$ can now be replaced by its expected value obtained from the probability distribution of $\btheta^*$, \ie
\begin{align*}
\matH &\approx \Ex_{\btheta^*}[\matH]\\
\numberthis \label{eq:H_approximation}
&=\ov{h} \matR
\end{align*}
with
\begin{align*}
\ov{h}
&= \Ex_{\btheta^*,\bx_k}[\mathcal{L}(\bx_k\T \btheta^* + \eta)]\\
&\approx \frac{\sqrt{\pi}}{2} \mcN(\eta;0,2v + 2)\\
\numberthis \label{mean-h_k}
&= \frac{1}{4} \frac{1}{\sqrt{v + 1}} \exp\left[\frac{-\eta^2}{4(v + 1)}\right]
\end{align*}
(for details, see \ref{Appendix.Expectation}). Next, substituting \eqref{covariance.matrix} and \eqref{eq:H_approximation} into \eqref{eq:weight-behavior}, and recalling the well-known property of the Elo algorithm which establishes that the sum of the skills does not change over time \cite[Ch.~5]{Langeville12_book} given the symmetrical nature of the ratings update and fixed set of players, the mean behavior of each individual skill can be expressed as
\begin{align*}
\Ex[\theta_{k+1,m}] &\approx \Ex[\theta_{k,m}] - \frac{2 \beta \ov{h}}{M-1} \left( \Ex[\theta_{k,m}] - \theta^*_m - \frac{1}{M} \sum_{i=1}^{M}
\Ex[\theta_{k,i} - \theta^*_i] \right)\\
\numberthis \label{eq:E[theta_k+1]}
&= \alpha_1 \Ex[\theta_{k,m}] + (1 - \alpha_1) \theta^*_m
\end{align*}
where
\begin{align} \label{eq:alpha_1}
\alpha_1 &= 1 - \frac{2 \beta \ov{h}}{M-1}
\end{align}
characterizes the convergence speed of the skills in the mean sense.

Therefore, since we assumed $\btheta_0 = \bzero$ in \eqref{eq:rating-sgd2}, \eqref{eq:E[theta_k+1]}  can be compactly written as 
\begin{align}
\label{eq:skill-behavior}
\Ex[\theta_{k,m}] &=  (1 - \alpha_1^k)\theta^*_m
\end{align}
which converges to $\btheta^*$ as $k$ increases if $|\alpha_1|<1$, \ie
\begin{align}
\label{lim.k.Ex.theta}
    \lim_{k\rightarrow\infty} \Ex[\theta_{k,m}] &=  \theta^*_m.
\end{align}
Based on \eqref{eq:skill-behavior}, the evolution of the skills can be predicted (as illustrated in Section~\ref{Sec:experimental_results}), converging to the ``true skills" at an exponential rate \cite[Th.~2.2]{jabin2015}.

\subsection{Mean-square deviation of the skills}\label{Sec:skills.variance}

Even when \eqref{lim.k.Ex.theta} is satisfied, it is still important to assess the uncertainty related to the estimate of skills, \ie the random fluctuations around the mean value. To this end, we define the \gls{msd} of the skills as
\begin{align}\label{eq:skills.variance0}
\ov{d}_k = \Ex[\tilde\btheta_k\T\tilde\btheta_k]
\end{align}
where
\begin{align}
\label{tilde.btheta}
\tilde\btheta_k &= \btheta_k - \btheta^*
\end{align}
represents the estimation error of the skills. Then, rewriting \eqref{eq:sgd-taylor.2} in terms of $\tilde\btheta_k$ as
\begin{align}\label{eq:param-error}
\tilde\btheta_{k+1} &= 
(\matI - \beta h_k \bx_k\bx_k\T) \tilde\btheta_k + \beta g_k \bx_k
\end{align} 
determining the inner product $\tilde\btheta_k\T\tilde\btheta_k$, noting that $\bx_k\T\bx_k = 2$, and taking the expected value of both sides of the resulting expression, we get
\begin{equation}
\label{eq:skills.variance1}
\begin{aligned}
\ov{d}_{k+1} &=  \ov{d}_k - 2\beta  \Ex[ h_k \tilde\btheta_k\T \bx_k\bx_k\T \tilde\btheta_k ] + 2\beta^2 \Ex[ h_k^2 \tilde\btheta_k\T \bx_k  \bx_k\T \tilde\btheta_k] \\
& + 2 \beta  \Ex[ g_k \bx_k\T \tilde\btheta_k] - 4\beta^2 \Ex[ g_k h_k \bx_k\T \tilde\btheta_k ] + 2\beta^2 \Ex[ g_k^2].
\end{aligned}
\end{equation}

Now, using the same approach as in \eqref{eq:H_approximation}, we approximate the second and third terms of the \gls{rhs} of \eqref{eq:skills.variance1} as
\begin{align}\label{eq:expected_value1}
\Ex[ h_k \tilde\btheta_k\T \bx_k\bx_k\T \tilde\btheta_k ] \approx \ov{h} \Ex[ \tilde\btheta_k\T \matR \tilde\btheta_k ]
\end{align}
and
\begin{align}\label{eq:expected_value2}
\Ex[ h_k^2 \tilde\btheta_k\T \bx_k\bx_k\T \tilde\btheta_k ] \approx \ov{h^2} \Ex[ \tilde\btheta_k\T \matR \tilde\btheta_k ]
\end{align}
in which $\ov{h}$ is defined in \eqref{mean-h_k},
\begin{align*}\label{mean-h_k^2}
\ov{h^2} &= \Ex_{\btheta^*,\bx_k}[\mcL^2(\bx_k\T \btheta^* + \eta)]\\
&\approx \frac{\sqrt{2\pi}}{16} \mcN(\eta;0,2v + 1)\\
&= \frac{1}{16} \sqrt{\frac{1}{2v + 1}} \exp{\left[\frac{-\eta^2}{2 (2v + 1)}\right]} \numberthis
\end{align*}
(for details, see \ref{Appendix.Expectation}), while
\begin{align*}
\Ex[ \tilde\btheta_k\T \matR \tilde\btheta_k ] &= \frac{2}{M-1} \Ex[ \tilde\btheta_k\T \tilde\btheta_k ] - \frac{2}{M(M-1)} \Ex[ \tilde\btheta_k\T \bone\bone\T \tilde\btheta_k ]\\
\label{eq:theta_R_theta}
&= \frac{2}{M-1} \ov{d}_k \numberthis
\end{align*}
since $\bone\T \tilde\btheta_k = 0$. Next, invoking the orthogonality principle \cite[Sec.~3.3]{farhang2013}, it is possible to show that the fourth and fifth terms of the \gls{rhs} of \eqref{eq:skills.variance1} reduce to
\begin{align}
\Ex[ g_k h_k \bx_k\T \tilde\btheta_k ] &= 0
\end{align}
and
\begin{align} \label{eq:ortho-g_k}
\Ex[ g_k \bx_k\T \tilde\btheta_k] &= 0.
\end{align}
While, the last term of the \gls{rhs} of \eqref{eq:skills.variance1} is derived from \eqref{eq:gradient-optimal} as
\begin{align*}
\Ex[ g_k^2 ] &= \Ex_{\btheta^*,\bx_k,y_k}\{ [y_k - \sigma(\bx_k\T\btheta^* + \eta)]^2 \} \\
&= \Ex_{\btheta^*,\bx_k,y_k}[y_k^2 -2y_k\sigma(\bx_k\T\btheta^* + \eta) + \sigma^2(\bx_k\T\btheta^* + \eta) ] \\
&= \Ex_{\btheta^*,\bx_k}\{ \sigma(\bx_k\T\btheta^* + \eta)[1 - \sigma(\bx_k\T\btheta^* + \eta)] \}\\
&= \Ex_{\btheta^*,\bx_k}[\mcL(\bx_k\T \btheta^* + \eta)]\\
&= \ov{h}. \numberthis \label{eq:expected_value_gk}
\end{align*}

Finally, substituting \eqref{eq:expected_value1}-\eqref{eq:expected_value_gk} into \eqref{eq:skills.variance1}, we obtain
\begin{align}\label{eq:skills.variance4}
\ov{d}_{k+1} &= \alpha_2\ov{d}_k + 2\beta^2 \ov{h}
\end{align}
where
\begin{align}\label{eq:alpha_2}
\alpha_2 &= 1 - \frac{4}{M-1} \beta (\ov{h} - \beta \ov{h^2})
\end{align}
represents the convergence speed in the mean-square sense. Alternatively, when $|\alpha_2| < 1$, \eqref{eq:skills.variance4} can be rewritten in a more convenient form as
\begin{align}
\label{eq:v_k-final}
\ov{d}_k = \alpha_2^k (\ov{d}_0 - \ov{d}_{\infty}) + \ov{d}_{\infty}
\end{align}
where
\begin{align} \label{eq:msd_0}
\ov{d}_0 &= M v
\end{align}
due to the initialization considered in \eqref{eq:rating-sgd2} and
\begin{align*}\label{eq:msd_inf}
\ov{d}_{\infty} &= \frac{2 \beta^2 \ov{h}}{1 - \alpha_2}\\
\numberthis
&= \frac{\beta \ov{h} (M-1)}{2(\ov{h} - \beta \ov{h^2})}
\end{align*}
comes from \eqref{eq:skills.variance4} by assuming that $d_\infty=d_{k+1}=d_k$. Therefore, from \eqref{eq:v_k-final}, the \gls{msd} can be predicted (as illustrated in Section~\ref{Sec:experimental_results}).

\subsection{Variance of the skills} \label{Sec:Variance.skills}

To have a better understanding of the statistics of skills, let us use the well-known bias-variance decomposition (see \cite[Sec.~4.4]{aldous2017}) to rewrite the \gls{msd} [defined in \eqref{eq:skills.variance0}] as
\begin{align*}
\ov{d}_k &= \Ex[(\tilde\btheta_k -\Ex[\tilde\btheta_k])\T (\tilde\btheta_k -\Ex[\tilde\btheta_k])] +  \Ex[\tilde\btheta_k\T] \Ex[\tilde\btheta_k]\\
&= \ov\omega_k + \ov{b}_k \numberthis \label{bias.variance.decomposition}
\end{align*}
where
\begin{align}
\label{skills.bias}
\ov{b}_k &= \alpha_1^{2k} \ov{d}_0
\end{align}
is the squared estimation bias obtained from \eqref{eq:skill-behavior}, and
\begin{align}
\label{variance.skills}
\ov\omega_k = (\alpha_2^k - \alpha_1^{2k}) \ov{d}_0 + (1-\alpha_2^k)\ov{d}_{\infty} 
\end{align}
is the total variance (\ie the sum of all variances) derived from \eqref{eq:v_k-final} and \eqref{skills.bias}. Note from \eqref{bias.variance.decomposition}-\eqref{variance.skills} that the bias-variance trade-off is actually a function of $k$. In particular, for $k=0$, the total variance is null (\ie $\ov\omega_0=0$) and the squared estimation bias is at its maximum (\ie $\ov{b}_0 = \ov{d}_0$). On the other hand, as $k$ increases, the squared estimation bias tends to zero (\ie $\ov{b}_\infty = 0$) and the total variance reaches its maximum (\ie $\ov\omega_\infty = \ov{d}_\infty$).

\subsection{Mean behavior of the loss function}\label{Sec:loss.mean}

Since the loss function is our performance criterion, here we derive an expression characterizing its behavior in the mean sense. To this end, let us substitute \eqref{eq:gradient-optimal} and \eqref{eq:hessian-optimal2} into  \eqref{eq:cost-taylor}, take the expected value of both sides of the resulting expression, and recall \eqref{eq:ortho-g_k}, so that 
\begin{align*}
\ov\ell_k &= \Ex[\tilde\ell_k(\btheta_k)]\\
&= \Ex[\ell_k(\btheta^*)] + \frac{1}{2} \Ex[h_k \tilde\btheta_k\T \bx_k\bx_k\T \tilde\btheta_k].
\label{eq:loss-behavior0}\numberthis
\end{align*}
Next, using the ergodicity argument along with the derivations presented in \ref{Appendix.Expectation}, the first term of the \gls{rhs} of \eqref{eq:loss-behavior0} can be determined as
\begin{align*}\label{eq:l_min}
\ov\ell_{\rm min} &= \Ex_{\btheta^*,\bx_k,y_k}[\ell_k(\btheta^*)]\\
&= \Ex_{\btheta^*,\bx_k,y_k}\{-y_k \ln[\sigma(\bx_k\T \btheta^* + \eta)] - (1 - y_k) \ln[\sigma(-\bx_k\T \btheta^* + \eta)] \}\\
&= \Ex_{\btheta^*,\bx_k}\{ -\sigma(\bx_k\T \btheta^* + \eta)\ln[\sigma(\bx_k\T \btheta^* + \eta)] - \sigma(-\bx_k\T \btheta^* + \eta)\ln[\sigma(-\bx_k\T \btheta^* + \eta)] \}\\
&\approx 2\ln(2)\sqrt{2\ln(2)\pi}\ \mcN(\eta;0,2v + 4\ln(2))\\
&= \ln(2) \sqrt{\frac{2\ln(2)}{v + 2\ln(2)}} \exp{\left[\frac{-\eta^2}{4v + 8\ln(2)}\right]} \numberthis
\end{align*}
representing a lower bound on the mean of the loss function. In turn, the second term of the \gls{rhs} of \eqref{eq:loss-behavior0} can be derived from \eqref{eq:expected_value1} and \eqref{eq:theta_R_theta} as 
\begin{align}
\label{eq:l_ex} \nonumber
\ov\ell_{{\rm ex},k} &\approx \frac{\ov{h}}{2} \Ex[\tilde\btheta_k\T \matR \tilde\btheta_k]\\
&= \frac{\ov{h}}{M-1} \ov{d}_k
\end{align}
characterizing the mean excess-loss due to the \gls{sg} optimization algorithm, which is a linear function of $\ov{d}_k$. Therefore, using \eqref{eq:l_min} and \eqref{eq:l_ex}, the mean of the loss function can be predicted through
\begin{align}
\label{eq:loss-behavior}
\ov\ell_k &\approx \ov\ell_{\rm min} + \ov\ell_{{\rm ex},k}
\end{align}
(as we illustrate in Section~\ref{Sec:experimental_results}). 

\section{Discussion}\label{Sec:Discussion}

Based on the model expressions derived so far (see Section~\ref{Sec:analysis}), some insights into the behavior of the rating algorithm (with respect to the number of games $k$, variance $v$, \gls{hfa} $\eta$, and step size $\beta$) are discussed below, in order to provide useful design guidelines.

\subsection{Time constants}\label{sec:time}
For an intuitive interpretation of convergence, the concept of time constant associated with the exponential terms observed in \eqref{eq:skill-behavior} and \eqref{eq:v_k-final} can be used by rewriting the exponential term based on $\alpha_1$ in \eqref{eq:skill-behavior} as follows:
\begin{align}
\label{alpha.1.power.k}
\alpha_1^k &=\e^{-\frac{k}{\tau_1}}
\end{align}
from which the convergence of skills $\btheta_{k}$ in the mean sense is characterized by the time constant $\tau_1 = -[\ln( \alpha_1)]^{-1}$. Thus, from \eqref{eq:alpha_1}, we have 
\begin{align}
\label{tau.1}
\tau_1 &\approx \frac{M-1}{2 \beta\ov{h}}.
\end{align}
Similarly, a time constant $\tau_2= -[\ln(\alpha_2)]^{-1}$ associated with the convergence of $\btheta_{k}$ in the mean-square sense [see \eqref{eq:v_k-final}] can be determined as
\begin{align}
\label{tau.2}
\tau_2 &\approx \frac{M-1}{4 \beta(\ov{h} - \beta \ov{h^2})}.
\end{align}
Approximations used in \eqref{tau.1} and \eqref{tau.2} hold for values of $\alpha_1$ and $\alpha_2$ near 1.

\begin{figure}[t!]
\centering
\includegraphics[scale=1]{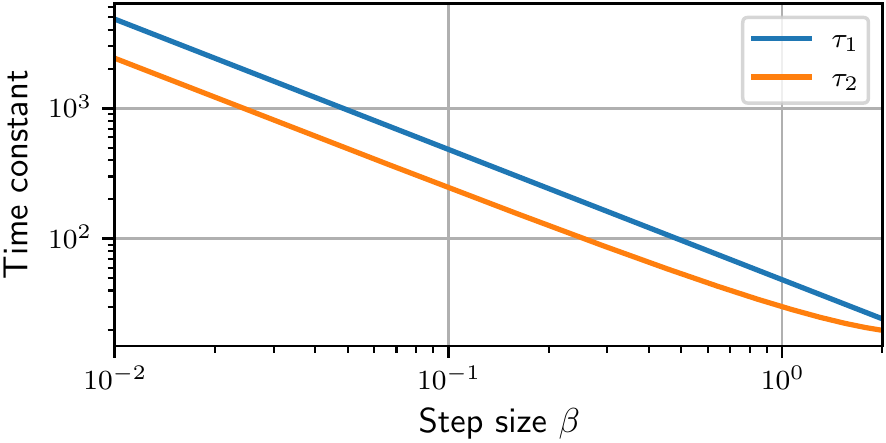}
\caption{Time constants related to the convergence of the rating algorithm in the mean $\tau_1$ [given by \eqref{tau.1}] and mean-square $\tau_2$ [given by \eqref{tau.2}] sense as a function of $\beta$, considering $M = 15$, $v=3$, and $\eta=0$.}
\label{fig:time_constant}
\end{figure}

Fig.~\ref{fig:time_constant} shows how the time constants $\tau_1$ [given by \eqref{tau.1}] and $\tau_2$ [given by \eqref{tau.2}] are related to the convergence speed of the rating algorithm in the mean and mean-square sense, respectively, as a function of the step-size value $\beta$, considering $M = 15$, $v=3$, and $\eta=0$. (Other values of $v$ are not considered since their impact on $\tau_1$ and $\tau_2$ is negligible.) Notice that these time constants are inversely proportional to the step size value; thus the convergence speed of the rating algorithm becomes slower as $\beta$ decreases and vice versa. Furthermore, observe that the algorithm presents different convergence requirements in terms of $\tau_1$ (associated with the estimate of skills $\btheta_k$) and $\tau_2$ (related to the uncertainty of the skill estimates $\ov{d}_k$); in particular, $\ov{d}_k$ decays faster than the estimated skills $\btheta_k$ improve for small $\beta$, \ie $\tau_1 \approx 2 \tau_2$ since the term $\beta\ov{h^2}$ in \eqref{tau.2} can be ignored. Besides the aforementioned aspects, this figure gives an idea that \eqref{tau.1} and \eqref{tau.2} may help in the selection of $\beta$ so that the algorithm reaches convergence within a predefined number of games.

\subsection{Meaning of ``convergence''}\label{sec:convergence}

At this point, it may be useful to differentiate between the concept of convergence understood as a mathematical condition that guarantees that $\btheta_k$ tends to $\btheta^*$ as $k$ increases and the one used colloquially\footnote{As pointed out critically in \cite[Sec.~3]{aldous2017}, the statement ``ratings tend to converge on a team’s true strength relative to its competitors after about 30 matches" is accepted at face value both in the literature and applications \cite{eloratings.net}, \cite{fide}. Note that the number of games is commonly viewed in a per-team basis; so, for comparison purposes, it is important to use $2k/M$ as the number of games instead of $k$.} in the rating literature that loosely assumes that convergence to the optimum is reached after an arbitrary and predefined number of games. However, this colloquial meaning does not hold because convergence cannot be characterized in a deterministic sense; in other words, it is not possible to establish any deterministic guarantee to reach the optimum within a finite number of games $k$. Therefore, the convergence of the rating algorithm can only be assessed in asymptotic or probabilistic sense.

Aiming to provide a practical guideline on the number of games required by the rating algorithm to reach convergence, we have to rely on some heuristics, such as:

\begin{itemize}

\item The approximation error $|\Ex[\theta_{k,m}]-\theta^*_m|$ becomes sufficiently small compared with the initial one $|\theta_{0,m}-\theta^*_m|$, which defines convergence in the mean sense; or

\item $\ov{d}_k$ (or, equivalently, $\ov{\ell}_k$) is sufficiently small compared with $\ov{d}_0$ (or, equivalently, $\ov{\ell}_0$), which defines convergence in the mean-square sense.

\end{itemize}

The ``sufficiently small" value can be predefined by specifying the number of games to reach convergence $K_\tr{c}$ as a multiple of the time constants $\tau_1$ or $\tau_2$. For example, for $K_\tr{c}=3\tau_1$, the exponential curve \eqref{alpha.1.power.k} decays 95\% of its initial value, becoming sufficiently flat to declare convergence by eyeballing. As a result, for $k>K_\tr{c}$, the approximation error $|\Ex[\theta_{k,m}]-\theta^*_m|$ becomes smaller than $5\%$ of the initial value $|\theta_{0,m}-\theta^*_m|$; also, since $\tau_1 \approx 2\tau_2$  for sufficiently small $\beta$, convergence in the mean sense implies that $\ov{d}_k$ practically equals its final value $\ov{d}_\infty$. So, by analyzing $\tau_1$ and $\tau_2$ we can define the step size $\beta$ to achieve convergence within a predefined number of games $k$.

\subsection{On the mean loss}\label{sec:loss}

Based on \eqref{eq:loss-behavior}, it can be verified that the mean loss $\ov\ell_{k}$ achieved by the algorithm is made up of a sum of two (non-negative) terms, namely $\ov\ell_{\tr{min}}$ [given by \eqref{eq:l_min}] and $\ov\ell_{{\rm ex},k}$ [given by \eqref{eq:l_ex}] which depend on $v$, $\eta$, $M$, $k$, and $\beta$. Particularly, $\ov\ell_{\tr{min}}$ depends only on the application/operating scenario considered (mainly through $v$), \ie it is not affected by the choice of step size $\beta$, \gls{hfa} $\eta$, number of teams $M$, and number of games $k$. In turn, $\beta$ and $k$ play important roles on the excess-loss $\ov\ell_{{\rm ex},k}$, increasing somehow the value of $\ov\ell_{k}$ achieved by the rating algorithm. So, it becomes helpful to clarify the impact of these variables on the performance of the rating algorithm.

\begin{figure}[t]
\centering
\includegraphics[scale=1]{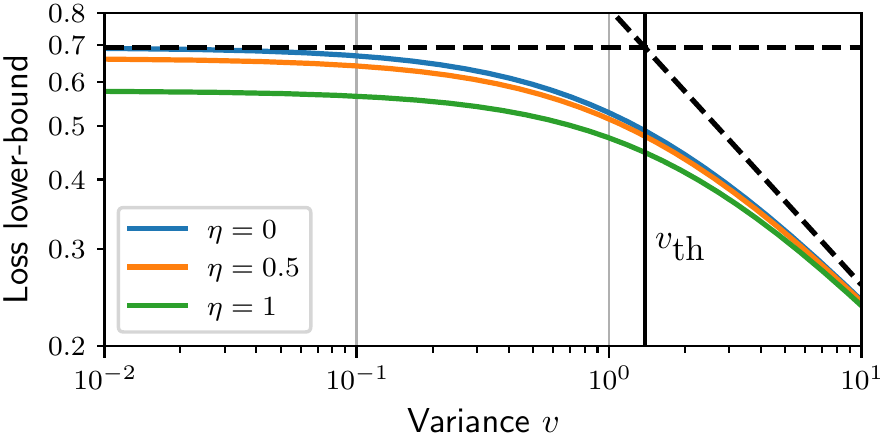}
\caption{Lower bound of the loss $\ov\ell_{\tr{min}}$ [given by \eqref{eq:l_min}] as a function of variance $v$, considering different values of $\eta$ and $M=15$. Asymptotes are shown by using dark-dashed lines and a vertical dark-solid line is used to indicate $v_{\rm th}$ [see \eqref{eq:vth}].}
\label{fig:loss_min}
\end{figure}

\figref{fig:loss_min} plots the lower-bound of the loss $\ov\ell_\tr{min}$ [given by \eqref{eq:l_min}] as a function of the variance $v$, considering different values of $\eta$ and $M=15$. This figure emphasizes that the performance of the algorithm degrades as $v$ decreases due to the high uncertainty of the results; which leads to larger values of $\ov\ell_\tr{min}$. On the other hand, when $v$ increases, the difference between the skills of the players may be large and the results of the games between these players are more easily predicted; hence, $\ov\ell_\tr{min}$ becomes smaller. Still, note that the impact of the \gls{hfa} parameter $\eta$ on $\ov\ell_\tr{min}$ is less significant as $v$ increases, since $\eta$ in \eqref{eq:l_min} can be ignored when the differences between skills are large.

In order to characterize the meaning of small and large $v$ (high and low uncertainty), let us approximate the curves in \figref{fig:loss_min} by using two asymptotes, one of them valid when $v\rightarrow 0$ and the other for $v\rightarrow\infty$. Thereby, we can determine a boundary/threshold $v_\tr{th}$ from the intersection of these two asymptotes [arising from \eqref{eq:l_min}], \ie
\begin{align}
    \ov{\ell}_\tr{min}\Big|_{v=0} &= \ov{\ell}_\tr{min}\Big|_{v\gg 2\ln(2)}
\end{align}
which results in
\begin{align*}\label{eq:vth}
    v_\tr{th} &= 2\ln(2)\\
              &\approx 1.4.\numberthis
\end{align*}
So, this value $v_\tr{th}$ establishes a boundary that separates small and large $v$.

\begin{figure}[t]
\centering
\includegraphics[scale=1]{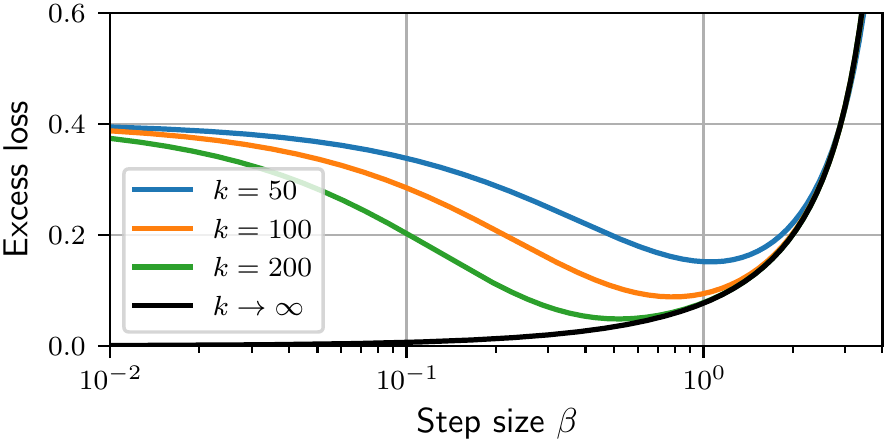}
\caption{Excess loss [given by \eqref{eq:l_ex}] as a function of the step-size value $\beta$, considering different number of games $k$ while $\eta = 0$, $v = 3$, and $M = 15$.}
\label{fig:loss_ex}
\end{figure}

\figref{fig:loss_ex} plots the mean excess-loss $\ov\ell_{{\rm ex},k}$ [given by \eqref{eq:l_ex}] as a function of $\beta$, considering different number of games $k$ while $\eta = 0$, $v = 3$, and $M = 15$. This figure ratifies the impact of $\beta$ and $k$ on the excess loss attained by the rating algorithm. Specifically, for a large number of games ($k\to\infty$), the excess loss exhibits a monotonically increasing characteristic with respect to $\beta$, \ie large $\beta$ yields higher values of $\ov\ell_{\tr{ex},k}$. On the other hand, for finite $k$ (\eg 50, 100, or 200 games), a unimodal characteristic (\ie unique minimum) is observed in the mean excess-loss\footnote{The loss $\ov\ell_k$ [given by \eqref{eq:loss-behavior}] will also exhibit the same convex characteristic.} due to the bias-variance trade-off (see Section~\ref{Sec:Variance.skills}); so, there is an optimum step-size value $\beta_{\tr{o}, k}$ which leads to the minimum attainable value of $\ov\ell_{\tr{ex},k}$ for a given number of games $k$ (as discussed later in Section~\ref{sec:optimum_step}). This characteristic is due to the fact that the rating algorithm is not capable of converging within $k$ games for small values of $\beta$; as a consequence, $\ov\ell_{\tr{ex},k}$ may be higher than the one achieved with intermediate values of $\beta$. Therefore, since $k$ is generally finite, choosing an appropriate value for $\beta$ is of paramount importance.

\subsection{Improvement over initialization}
A condition on the step-size $\beta$ is now derived to ensure that the uncertainty about the estimated skills $\btheta_k$ (characterized by $\ov{d}_k$) is reduced over its initial value as $k$ increases, \ie
\begin{align}\label{ineq:improvement_msd}
\ov{d}_k &< \ov{d}_0.
\end{align}
To this end, substituting \eqref{eq:v_k-final} into \eqref{ineq:improvement_msd}, we have
\begin{align}
\label{ineq:improvement_msd3}
(1 - \alpha_2^k) (\ov{d}_0 - \ov{d}_{\infty}) &> 0.
\end{align}
Then, since $|\alpha_2| < 1$ implies that $(1-\alpha_2^k)$ is positive, 
\eqref{ineq:improvement_msd3} reduces to\footnote{
Note that \eqref{ineq:improvement_msd4} can be reinterpreted in terms of the bias-variance trade-off as $\ov\omega_\infty < \ov{b}_0$; so improvement occurs when the maximum variance $\ov\omega_\infty$ is smaller than the initial squared bias $\ov{b}_0$.}
\begin{align}
\label{ineq:improvement_msd4}
\ov{d}_\infty &< \ov{d}_0.
\end{align}
Next, using \eqref{eq:msd_0} and \eqref{eq:msd_inf} in \eqref{ineq:improvement_msd4}, we get
\begin{align}
\frac{\beta \ov{h} (M-1)}{2(\ov{h} - \beta \ov{h^2})} &< M v
\end{align}
and solving the resulting expression for $\beta$, leads to
\begin{align}
\label{ineq:improvement.condition}
0 < \beta < \left[ \frac{(1 - 1/M)}{2v} + \frac{\ov{h^2}}{\ov{h}} \right]^{-1}.
\end{align}
Note that, assuming $M$ large, \eqref{ineq:improvement.condition} can be asymptotically approximated as
\begin{align}
0 < \beta < \left( \frac{1}{2v} + \frac{\ov{h^2}}{\ov{h}} \right)^{-1}
\end{align}
and, given that $\ov{h^2}/\ov{h}$ is limited to $[\sqrt{2}/8, 1/4]$ such that it can be ignored for small $v$, we have\footnote{In practice, $\beta$ is typically chosen smaller than \eqref{eq:improvement_over_small_v}. For example, \gls{fifa} \cite{fifa_ranking} uses $\beta\in[0.02,0.23]$ in football rankings and \gls{fide} \cite{fide} considers $\beta\in[0.06,0.23]$ in chess rankings. Note that these values have been obtained by scaling the step-size value $\beta'$ provided by either \gls{fifa} or \gls{fide} through $\beta = \beta'/s'$, with $s'$ defined in \eqref{scale-ln10}.}
\begin{align}\label{eq:improvement_over_small_v}
0 < \beta < 2v.
\end{align}

\begin{figure}[t!]
\centering
\includegraphics[scale=1]{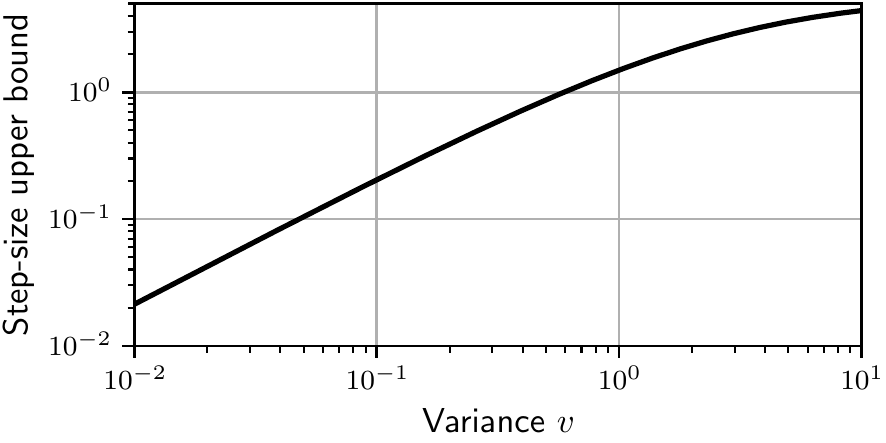}\\
\caption{Condition on $\beta$ for improvement over initialization [given by \eqref{ineq:improvement.condition}] as a function of $v$, considering $M = 15$ and $\eta = 0$.}
\label{fig:improvement_over}
\end{figure}

\figref{fig:improvement_over} plots the upper bound on the  step-size value $\beta$ given in \eqref{ineq:improvement.condition} as a function of $v$, considering $M = 15$ and $\eta = 0$. Other values of $M$ and $\eta$ have not been used, as they do not significantly affect \eqref{ineq:improvement.condition}. Notice that the upper bound on $\beta$ is lower as $v$ decreases, exhibiting an almost linear dependence on $v$; hence, \eqref{eq:improvement_over_small_v} holds as a rule of thumb. Nevertheless, we emphasize that such a bound may be too loose and merely guarantees the improvement over initialization; therefore, in practice, it may be preferable to choose a $\beta$ smaller than \eqref{eq:improvement_over_small_v} especially when the number of games $k$ is limited.

\subsection{Optimum step-size value}\label{sec:optimum_step}

\begin{figure}[t]
\centering
\includegraphics[scale=1]{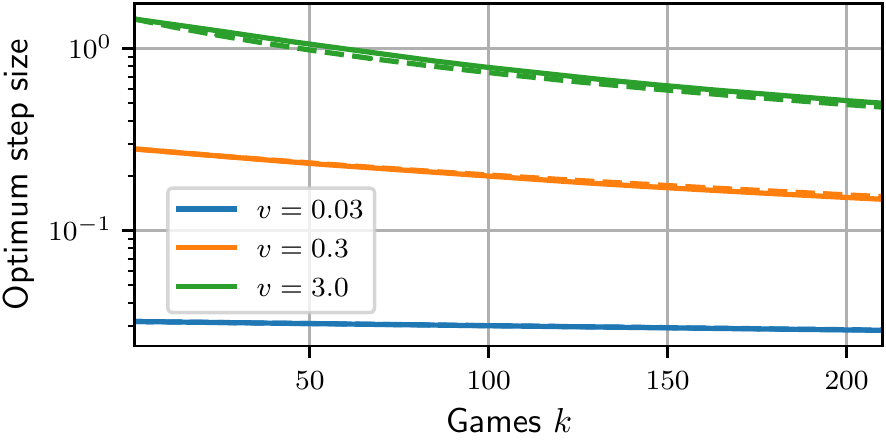}\\
\caption{Optimum step-size value $\beta_{{\tr{o}},k}$ obtained from the numerical solution of \eqref{eq:optimum-beta} (solid lines) and the approximate solution \eqref{eq:optimum_step} (dashed lines) as a function of the number of games $k$, considering different values of $v$ with $M = 15$ and $\eta = 0$. Curves for other values of $\eta$ are omitted since this variable does not significantly affect \eqref{eq:optimum_step}.}
\label{fig:optimal_beta}
\end{figure}

An optimum value for step-size $\beta$, considering a particular number of games $k$, can be determined to minimize \eqref{eq:v_k-final} [or, equivalently, the mean loss \eqref{eq:loss-behavior}]; specifically, we need to find a solution for the following optimization problem:
\begin{equation}\label{eq:optimum-beta}
\beta_{{\rm o}, k} = \argmin_{\beta}\; \ov{d}_k.
\end{equation}
Since a closed-form solution for \eqref{eq:optimum-beta} cannot be straightforwardly obtained for an arbitrary $k$, we resort to approximations (shown in \ref{Appendix.Step}) to obtain
\begin{align}\label{eq:optimum_step}
\beta_{\tr{o},k} &\approx \frac{1}{2} \left[ \frac{(1 - 1/M)}{2v} + \frac{\ov{h^2}}{\ov{h}} + \frac{2 \ov{h^2} (k - 1)}{M - 1} \right]^{-1}.
\end{align}

\figref{fig:optimal_beta} plots the optimal step-size value $\beta_{\tr{o},k}$ as a function of the number of games $k$, for  different values of $v$ with $M=15$ and $\eta=0$. We consider both the numerical solution of \eqref{eq:optimum-beta} and the approximate solution \eqref{eq:optimum_step}, and observe that 

\begin{itemize}
\item The results calculated from \eqref{eq:optimum_step} match very well with those determined from the numerical solution of \eqref{eq:optimum-beta} regardless of $v$ and $k$, thus confirming the practicality of the approximate solution obtained.

\item The optimal step size $\beta_{\tr{o},k}$ exhibits a decreasing characteristic with respect to the number of games $k$, \ie larger step sizes are needed to improve the convergence of the rating algorithm when $k$ is small. 

\item The impact of $v$ on the choice of the step-size value is more significant than the number of games $k$; in particular, for small $v$, $k$ can be neglected, and we can use $\beta_{\tr{o},k}\approx v$.

\end{itemize}
\noindent
Therefore, the choice of the step-size value $\beta$ should be made carefully depending on the competition scenario.

\section{Application on real-world data}\label{Sec:experimental_results}

Aiming to assess the accuracy of the expressions derived from our analysis, the results obtained by running the rating algorithm [given in \eqref{eq:rating-sgd2}] are compared to the behavior described through the proposed model\footnote{The code used in the experiments is available at \url{https://github.com/dangpzanco/elo-rating}.}. To this end, we use real-world data from the Italian professional volleyball league (SuperLega), spanning 10 seasons from 2009/10 to 2018/19 because of its completeness\footnote{Seasons 2019/20, 2020/21, and 2021/22 were affected by the pandemic, while season 2022/23 was intentionally omitted to prevent incorporating unanalyzed data.}, whose results are available on FlashScore's website \cite{flashscore.ca}. We consider only the games in a regular season and neglect the playoffs; by doing so, we guarantee that all teams play the same number of games against each other, and, hence, modeling the scheduling as uniform is appropriate [\ie the regular season is a round-robin tournament with empirical frequencies equal to the theoretical ones given in \eqref{eq:PR.schedules} and \eqref{eq:PR.schedules.cond}]. Note that the aim is to verify whether our analysis characterizes well the behavior of the rating algorithm for a known operating scenario, thus ratifying the insights discussed from our analysis.

The \gls{hfa} parameter $\eta$ (used in the rating algorithm) and the variance of the skills $v$ (required in the stochastic model) are assumed known a priori in each season so that $\ov{h}$ [see \eqref{mean-h_k}], $\ov{h^2}$ [see \eqref{mean-h_k^2}], and $\ov\ell_{\rm min}$ [see \eqref{eq:l_min}] can be properly computed to validate the analysis. Specifically, $\eta$ may be treated as a parameter along with $\btheta$ in \eqref{ML.optimization} and, hence, both of them can be estimated by using any popular machine-learning approach \cite{goodfellow-deeplearning}, yielding thus $\hat\eta$ and $\hat\btheta$; once $\hat\btheta$ is obtained, we can determine $\hat{v}$ directly from its (unbiased) sample variance, since the scheduling is uniform. Note that we have used here the \gls{sg} optimization method with a very small step size and a large number of epochs (passages through the same data).

A summary of \mbox{SuperLega}'s data is presented in \tabref{tab:summary}, highlighting the number of teams $M$ and games $K$ along with the estimated values of the \gls{hfa} parameter $\eta$ and the variance of skills $v$. Notice that $M$ varies (within the range of 12 to 15); as a consequence, $K$ changes (from 132 to 210 games) depending on the considered season. 
Note still that the \gls{hfa} parameter could be neglected here on both the algorithm and model, since the estimated variance $\hat{v}$ is relatively large which implies that the difference between the skills is more significant than the additional change due to $\hat{\eta}$.

\begin{table}[t!]
\centering
\caption{Summary of the SuperLega's dataset.}
\label{tab:summary}
\medskip
\begin{tabular}{@{}ccccc@{}}
\toprule
Season    & $M$ & $K$ &  $\hat\eta$  &   $\hat{v}$    \\ \midrule
2009 & 15 & 210 & 0.66 & 2.7 \\
2010 & 14 & 182 & 0.32 & 1.6 \\
2011 & 14 & 182 & 0.35 & 1.2 \\
2012 & 12 & 132 & 0.40 & 1.9 \\
2013 & 12 & 132 & 0.55 & 1.3 \\
2014 & 13 & 156 & 0.47 & 2.9 \\
2015 & 12 & 132 & 0.06 & 2.4 \\
2016 & 14 & 182 & 0.77 & 2.4 \\
2017 & 14 & 182 & 0.22 & 3.0 \\
2018 & 14 & 182 & 0.49 & 3.7 \\ 
\bottomrule
\end{tabular}
\end{table}

\subsection{Example 1}

This example aims i) to assess the ability of the proposed model to characterize the behavior of the rating algorithm; and ii) to show how $\beta$ affects the \gls{msd} of the skills \eqref{eq:v_k-final} and the mean loss function \eqref{eq:loss-behavior}. To this end, different step-size values are used, namely: $\beta=0.1$ (within the range suggested by \cite{fifa_ranking,fide}), $\beta=0.87$ [obtained by averaging the optimum step-size value \eqref{eq:optimum_step} with $k=K/4$ for all seasons], and $\beta=2.49$ [determined by averaging the condition to improve over initialization \eqref{ineq:improvement.condition} for all seasons]. Note that the rating algorithm is run for each season and the mean behavior is computed (over the seasons) by truncating the number of games to the shortest season ($K=132$).

\begin{figure}[t]
\centering
\subfloat[]{\includegraphics[width=0.48\textwidth]{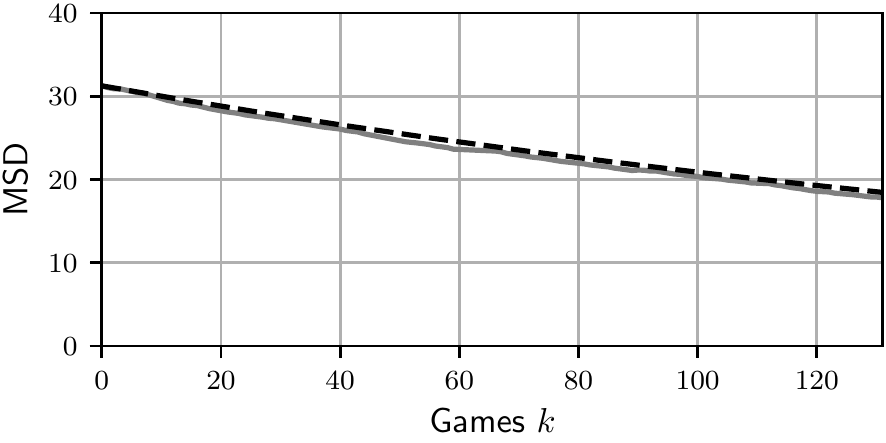}}
\quad
\subfloat[]{\includegraphics[width=0.48\textwidth]{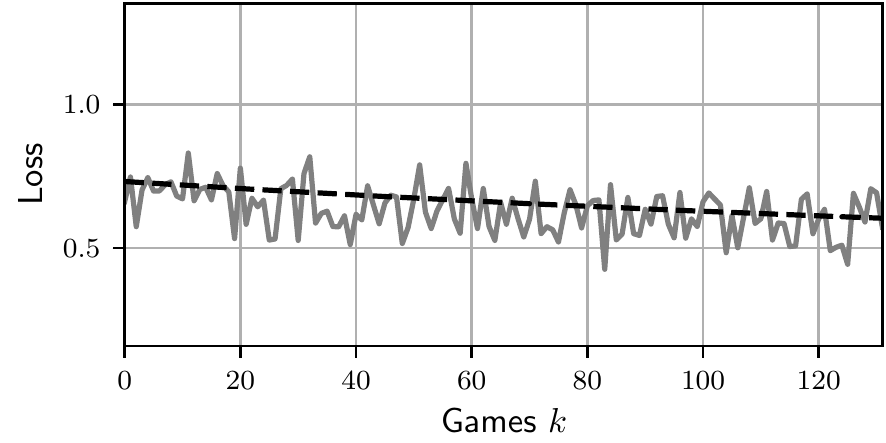}}\\
\subfloat[]{\includegraphics[width=0.48\textwidth]{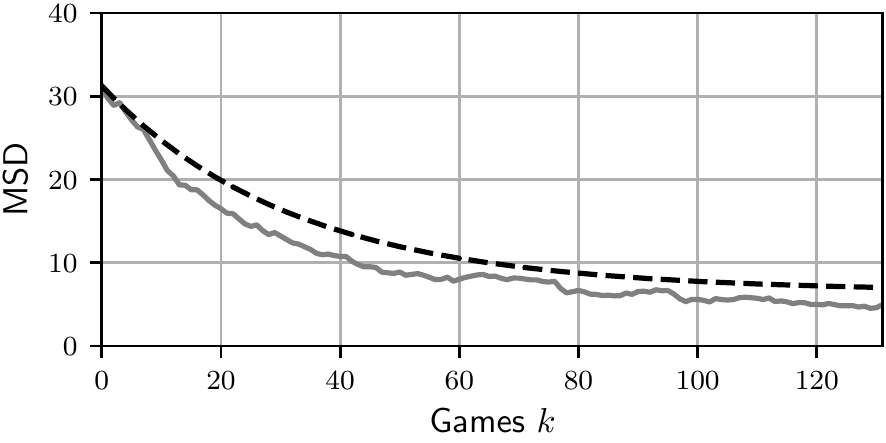}}
\quad
\subfloat[]{\includegraphics[width=0.48\textwidth]{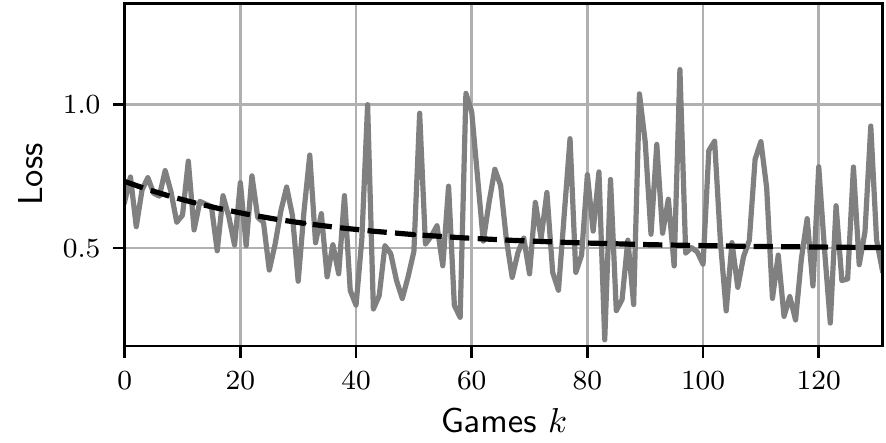}}\\
\subfloat[]{\includegraphics[width=0.48\textwidth]{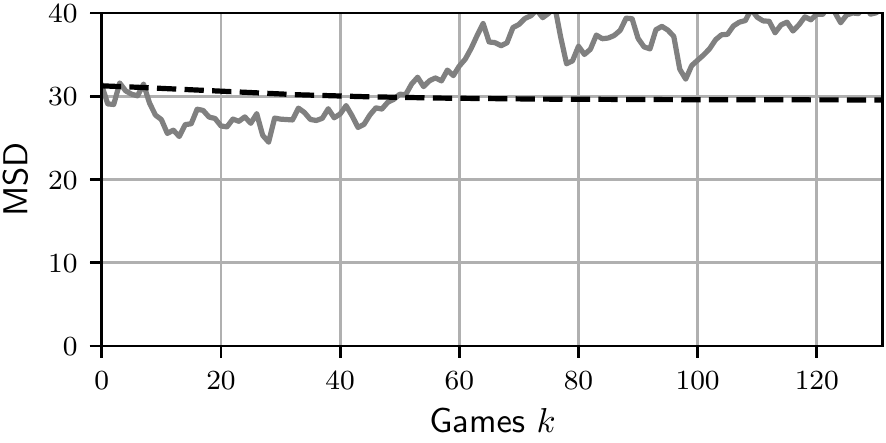}}
\quad
\subfloat[]{\includegraphics[width=0.48\textwidth]{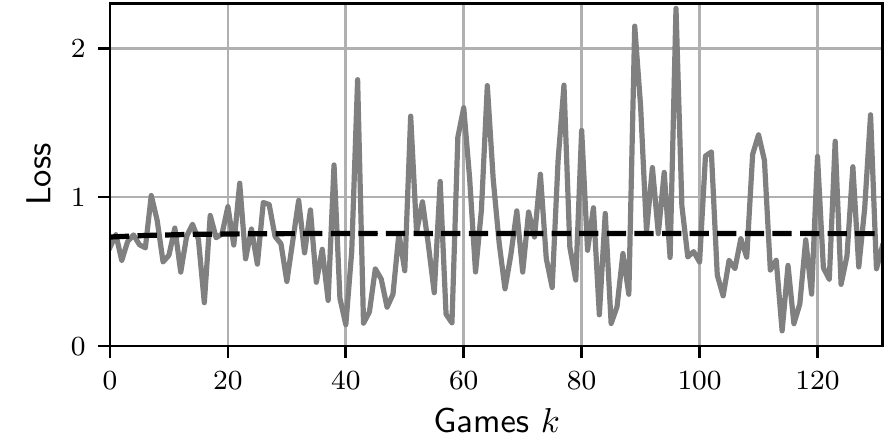}}\\
\caption{Example 1. Evolution of the \gls{msd} of the skills (left) and the mean loss function (right) obtained from simulation results (gray-solid lines) and proposed model (dark-dashed lines), considering (a)~and~(b)~$\beta = 0.1$, (c)~and~(d)~$\beta = 0.87$, and (e)~and~(f) $\beta = 2.49$.
}
\label{fig:exp_metrics}
\end{figure}

\figref{fig:exp_metrics} presents the evolution of the \gls{msd} of the skills and the mean loss function, considering $\beta=0.1$ [Figs.~\ref{fig:exp_metrics}(a)~and~(b)], $\beta=0.87$ [Figs.~\ref{fig:exp_metrics}(c)~and~(d)], and $\beta=2.49$ [Figs.~\ref{fig:exp_metrics}(e)~and~(f)]. Notice from these figures that the proposed model describes very well the behavior of the rating algorithm in the transient- and steady-state phases, irrespective of the considered step-size values. In addition to this aspect, from Figs.~\ref{fig:exp_metrics}(a)~and~(b), one can observe that the rating algorithm is not capable of reaching convergence until the end of the season for small values of $\beta$ (as suggested by \cite{fifa_ranking,fide}). In turn, when $\beta$ is adjusted according to the optimal step-size value [given by \eqref{eq:optimum_step}] with $k$ equal to a quarter of the season (\ie $k=K/4$ games), it can be verified that the convergence speed of the rating algorithm is significantly improved [as shown in Figs.~\ref{fig:exp_metrics}(c)~and~(d)]; in this condition, the algorithm reaches the optimal solution within $K$ games. Lastly, if we increase $\beta$ to the threshold derived as improvement over initialization [defined by \eqref{ineq:improvement.condition}], little to no improvement is observed as $k$ increases [as depicted in Figs.~\ref{fig:exp_metrics}(e)~and~(f)]. Therefore, we verify that the convergence of the rating algorithm is affected by the choice of $\beta$, ratifying the relevance of \eqref{eq:optimum_step}.

\subsection{Example 2}
In this example, the main goals are i) to assess the accuracy of the model expressions derived characterizing the \gls{msd} \eqref{eq:v_k-final} and the mean loss  \eqref{eq:loss-behavior} when $k$ increases at the end of the season; ii) to investigate the impact of step-size $\beta$ on the \gls{msd} and mean loss; and iii) to verify the practical applicability of the design guidelines provided about the choice of $\beta$ (as discussed in Section~\ref{Sec:Discussion}). To this end, we consider different values for the step-size, \ie $\beta\in(0.01, 4)$. Note that the rating algorithm is run for each season and the mean behavior is computed (over the seasons) by truncating the number of games to the shortest season ($K=132$) and averaging the last 10 samples of the variable of interest, in order to provide a better visualization of the experimental results.

\begin{figure}[t]
\centering
\subfloat[]{\includegraphics[width=0.48\textwidth]{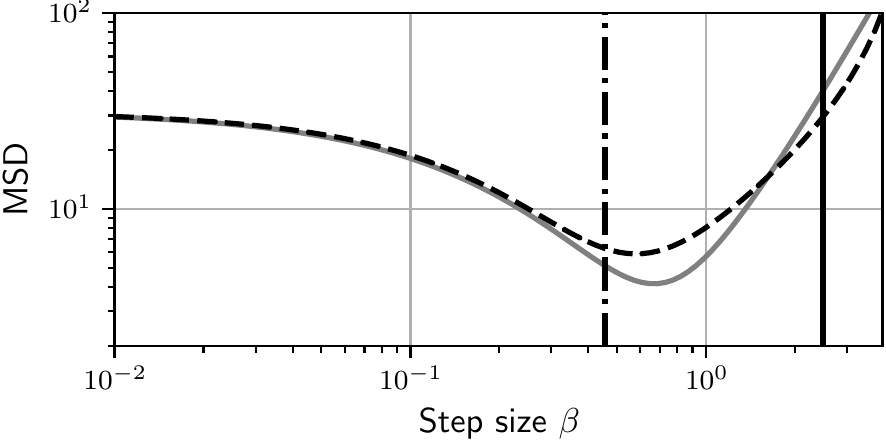}}
\quad
\subfloat[]{\includegraphics[width=0.48\textwidth]{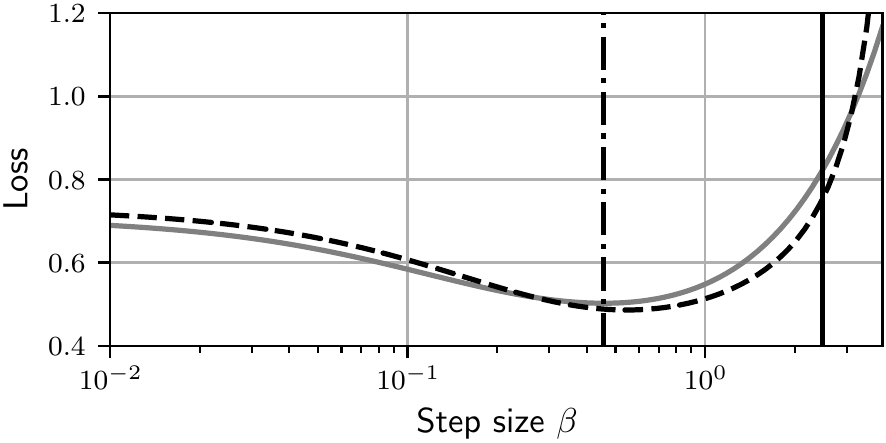}}\\
\caption{Example 2. Values of (a) \gls{msd} and (b) mean loss achieved by the rating algorithm as $k$ approaches $K$ obtained from simulation results (gray-solid lines) and proposed model (dark-dashed lines), considering different step-size values $\beta$. Vertical lines indicate the upper bound of the improvement over initialization condition \eqref{ineq:improvement.condition} (dark-solid line) and the optimum step-size value \eqref{eq:optimum-beta} for $k=K$ (dark dash-dotted line).}
\label{fig:exp-steady}
\end{figure}

\figref{fig:exp-steady} depicts the results obtained for the \gls{msd} [\figref{fig:exp-steady}(a)] and mean loss [\figref{fig:exp-steady}(b)] as a function of $\beta$. Based on this figure, one can draw the following conclusions:
\begin{itemize}
\item The proposed model describes very well the behavior of the rating algorithm (when approaching the end of the season) over a wide range step-size values $\beta$.  
\item Step-size values exceeding ``improvement over initialization'' [given  by \eqref{ineq:improvement.condition}] imply a high uncertainty of the skill estimates and predictions (\gls{msd} and mean loss, respectively); so, such values of $\beta$ should not be used.
\item Adjusting $\beta$ according to the optimal step-size value [given by \eqref{eq:optimum_step}] allows achieving a value of \gls{msd} and mean loss very close to their minimum within a certain number of games $k$.
\end{itemize}
Therefore, the design guidelines provided in Section~\ref{Sec:Discussion} also hold on empirical data and should be taken into account in practice.

\subsection{Example 3}

Our goal here is to show that the derived stochastic model explains well the evolution of the estimate of the skills $\btheta_k$ [through \eqref{eq:skill-behavior}] and \gls{msd} of the skills [given in \eqref{eq:v_k-final}]. Also, we aim to verify the impact of the data from different seasons on the ability of the derived model to describe the behavior of the rating algorithm. Specifically, the algorithm is evaluated using data from four distinct seasons, namely: 2009/10, 2010/11, 2015/16, and 2018/2019, chosen to consider diverse values of $\hat{v}$ and $\hat\eta$ (see Table~\ref{tab:summary}). Note that, instead of analyzing the averages over the seasons, the results of a single run of the algorithm (which is a realization of a random process) are compared with its mean behavior described from the model. Lastly, aiming to maintain methodological consistency, the optimum step-size value is computed from \eqref{eq:optimum_step} with $k=K/4$ for each selected season.

\begin{figure}[hb!]
\centering
\subfloat[]{\includegraphics[width=0.48\textwidth]{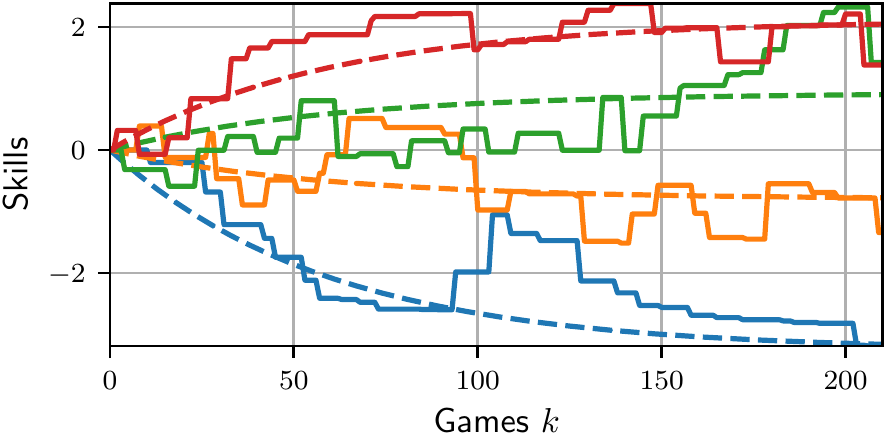}}
\quad
\subfloat[]{\includegraphics[width=0.48\textwidth]{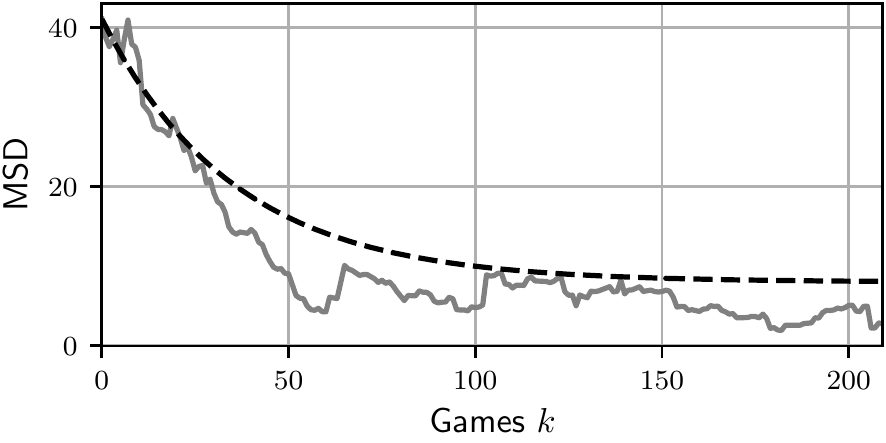}}\\
\subfloat[]{\includegraphics[width=0.48\textwidth]{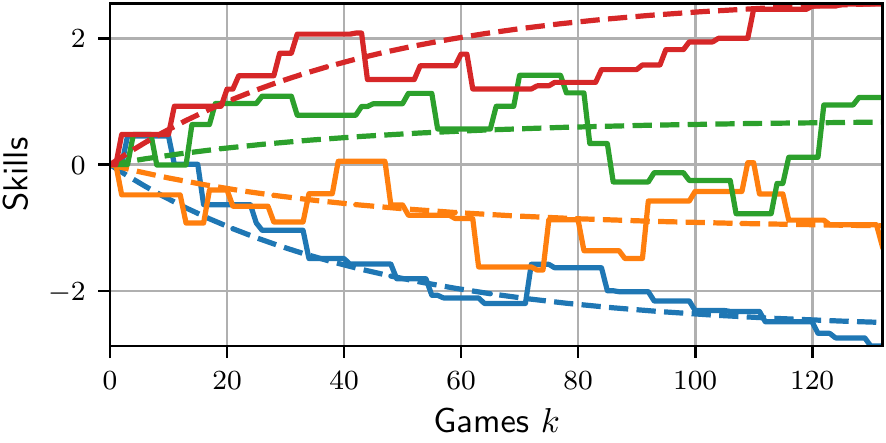}}
\quad
\subfloat[]{\includegraphics[width=0.48\textwidth]{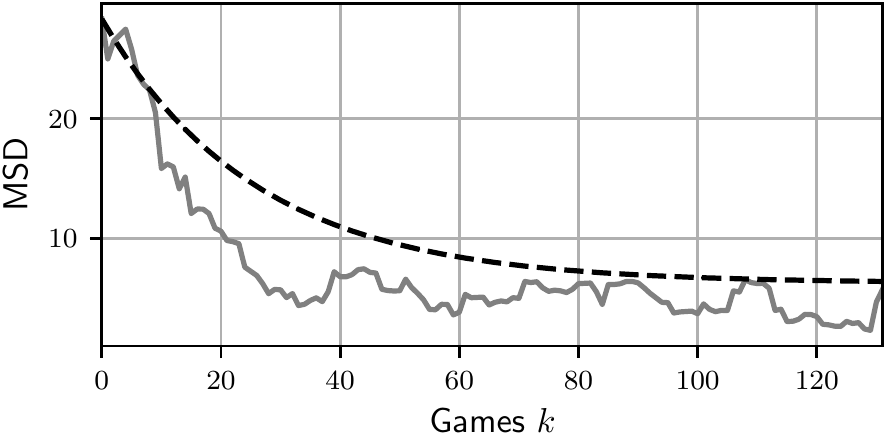}}\\
\subfloat[]{\includegraphics[width=0.48\textwidth]{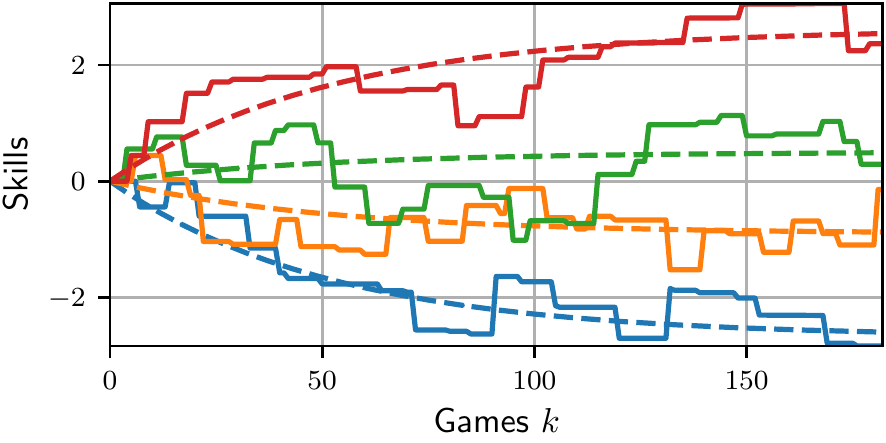}}
\quad
\subfloat[]{\includegraphics[width=0.48\textwidth]{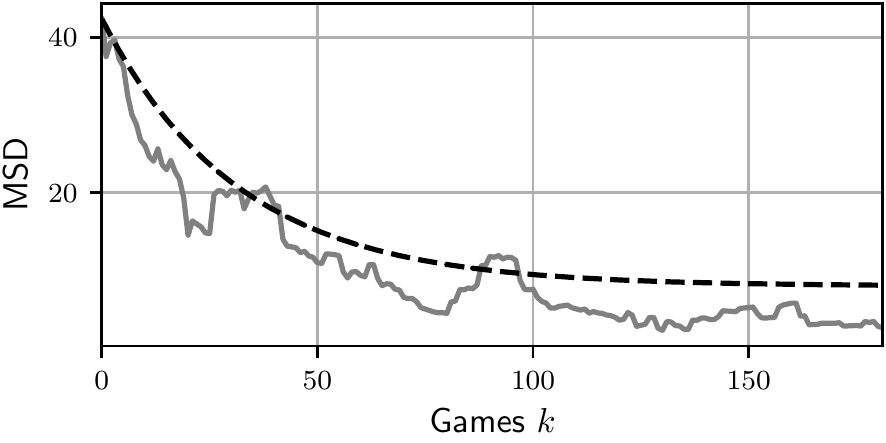}}\\
\subfloat[]{\includegraphics[width=0.48\textwidth]{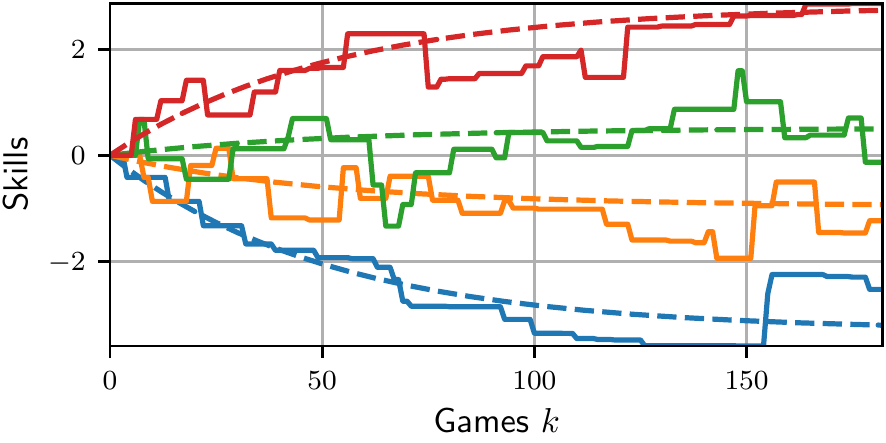}}
\quad
\subfloat[]{\includegraphics[width=0.48\textwidth]{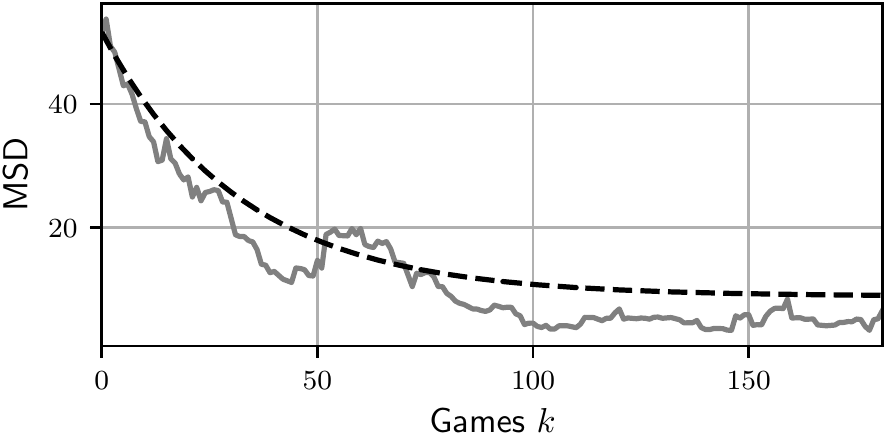}}
\caption{Example 3. Evolution of the skills for a subset of 4 teams (left) and \gls{msd} of the skills (right) obtained from simulation results (solid lines) and proposed model (dashed lines), considering season (a)~and~(b) 2009/10 with $\beta_{\tr{o},k}=0.939$, (c)~and~(d) 2015/16 with $\beta_{\tr{o},k}=0.998$, (e)~and~(f) 2017/18 with $\beta_{\tr{o},k}=0.93$, and (f)~and~(g) 2018/19 with $\beta_{\tr{o},k}=1.09$.} 
\label{fig:exp_seasons}
\end{figure}

\figref{fig:exp_seasons} presents the evolution of the skills (for four selected teams with distinct characteristics, thus yielding non-overlapping curves at the end of the season for better visualization purposes) and the corresponding \gls{msd} curves. Specifically, results for seasons 2009/10 [Fig.~\ref{fig:exp_seasons}(a)~and~\ref{fig:exp_seasons}(b)], 2015/16 [Fig.~\ref{fig:exp_seasons}(c)~and~\ref{fig:exp_seasons}(d)], 2017/18 [Fig.~\ref{fig:exp_seasons}(e)~ and~\ref{fig:exp_seasons}(f)], and 2018/19 [Fig.~\ref{fig:exp_seasons}(g)~and~\ref{fig:exp_seasons}(h)] are shown. Notice that the evolution of the skills obtained by running the algorithm has a staircase-like form\footnote{When averaging over different realizations of the scheduling, the staircase form would be replaced by the exponential form of the mean of skills $\Ex[\theta_m]$.} due to a particular realization of the scheduling, \ie the skills remain constant when the team is not playing. Observe also that, while the estimates $\btheta_k$ are random (owing to random outcome and scheduling of games), the analytical expressions derived for the mean of skills $\Ex[\btheta_k]$ as well as their \gls{msd} become a practical proxy for $\btheta_k$ and $\ov{d}_k$ regardless of the considered season. Therefore, we conclude that the proposed model characterizes well the behavior of the rating algorithm under a wide range of conditions, which validates the applicability of the design guidelines discussed so far (given in Section~\ref{Sec:Discussion}).


\section{Concluding remarks}\label{Sec:Conclusions}
In this paper, we presented a stochastic analysis of the well-known Elo algorithm, considering a round-robin tournament. Specifically, we derived analytical expressions to characterize the behavior/evolution of skills throughout a tournament. By using these expressions, the impact of the operating scenario and hyperparameters on the performance of the algorithm was discussed, with the aim of providing useful insight and design guidelines. Simulation results are shown by using real-world data from the Italian Volleyball League (SuperLega), confirming the applicability of the theoretical findings. 

The main take-home message and guidelines for the rating practitioners can be summarized as follows:  
\begin{itemize}   
    \item The convergence may be defined only in a probabilistic sense due to the randomness of the outcomes, scheduling, and skills. Consequently, the common belief of requiring approximately 30 games for convergence is misleading (see Section~\ref{sec:convergence}).
    
    \item The convergence in the mean and mean-square senses follows exponentially decreasing functions whose time constants are inversely proportional to the step-size parameter $\beta$ (see Section~\ref{sec:time}).

    \item The convergence is intrinsically related to the choice of the step-size value $\beta$ (see Section~\ref{sec:loss}), becoming generally slower as $\beta$ decreases (which implies a smaller variance of the estimation) and faster as $\beta$ increases (at the cost of higher variance of the estimation).
    
    \item The appropriate choice of $\beta$ becomes especially important given a finite number of games $K$. In particular, an intermediate (\ie optimum) value of $\beta$ should be used to improve convergence, since it may not be attained by using small values of $\beta$ (see Section~\ref{sec:optimum_step}).

    \item The predictive capacity of the algorithm is strongly affected by the variance of the skills $v$ (see Section~\ref{sec:loss}). In particular, for small variance, \eg $v<0.1$, the mean loss values at initialization and after adaptation are almost the same. This means that, in leagues with high competitive balance, which are characterized by small differences between skills, it may be difficult to declare convergence by observing the prediction of the game results.
\end{itemize}
Further research work could address the development of adjustment rules for the step size of the algorithm given its impact on the algorithm performance, to proceed with the stochastic analysis of other extensions and variations of the (original) Elo algorithm, as well as to consider the inclusion of draws and multiple outcomes in the model.

\section*{CRediT authorship contribution statement}
\textbf{Daniel Gomes de Pinho Zanco:} Investigation, Methodology, Software, Writing - Original Draft, Writing - Review \& Editing. \textbf{Leszek Szczecinski:} Conceptualization, Methodology, Project Administration, Supervision, Validation, Writing - Review \& Editing. \textbf{Eduardo Vinicius Kuhn:} Investigation, Writing - Original Draft, Writing - Review \& Editing. \textbf{Rui Seara:} Funding Acquisition, Resources, Writing - Review \& Editing.

\section*{Declaration of competing interest}
The authors declare that they have no known competing financial interests or personal relationships that could have appeared to influence the work reported in this paper.

\section*{Acknowledgment}
This research work was supported in part by the Coordination for the Improvement of Higher Education Personnel (CAPES) and the Brazilian National Council for Scientific and Technological Development (CNPq) from Brazil and by the Emerging Leaders in the Americas Program (ELAP) from the Government of Canada.

The authors are thankful to the Handling Editor and the anonymous reviewers for their constructive comments and suggestions.

The authors express their gratitude to Dr. Samuel Charles Passaglia, from the Kavli Institute for the Physics and Mathematics of the Universe, University of Tokyo, Japan, for his valuable comments and suggestions on the early version of the manuscript.

\appendix

\section{\normalfont\it Determination of \eqref{mean-h_k}, \eqref{mean-h_k^2}, and \eqref{eq:l_min}}\label{Appendix.Expectation}

We show how the Laplace approximation \cite[Ch.~5]{rdpeng-advstatcomp-online} can be applied to calculate the expected values in \eqref{mean-h_k}, \eqref{mean-h_k^2}, and \eqref{eq:l_min}, which involve a random variable of the form $f(z)$, with $f(\,\cdot\,)$ being a strictly positive, twice-differentiable, and absolutely integrable function, while $z$ is a Gaussian random variable with mean $\mu_z$ and variance $v_z$. Thereby, using the definition of expected value \cite{Therrien1992DRS}, we can write
\begin{align*}\label{log[f(z)]}
\Ex[f(z)] &= \int f(z) \mcN(z;\mu_z,v_z) \dd z\\
          &= \int \e^{\ln[f(z)]} \mcN(z;\mu_z,v_z) \dd z.\numberthis
\end{align*}
Next, expanding $\ln[f(z)]$ through Taylor series around $z=0$, \eqref{log[f(z)]} can be approximated to
\begin{align}\label{taylor_log[f(z)]}
\Ex[f(z)] &\approx\int \e^{\ln[f(0)] + [f'(0)/f(0)] z + \frac{1}{2}\{ f''(0)/f(0)-[f'(0)/f(0)]^2 \}z^2} \mcN(z;\mu_z,v_z) \dd z.
\end{align}
Then, assuming that $f(z)$ reaches its maximum at $z=0$ such that $f'(0) = 0$, \eqref{taylor_log[f(z)]} reduces to
\begin{align}\label{taylor_simplified}
\Ex[f(z)] &=f(0) \int \e^{\frac{1}{2} [f''(0)/f(0)] z^2} \mcN(z;\mu_z,v_z) \dd z.
\end{align}
Alternatively, since the exponential term resembles Gaussian \gls{pdf}, we can rewrite \eqref{taylor_simplified} as
\begin{align}\label{prod.Gaussian}
\Ex[f(z)] &=f(0) \sqrt{2\pi v_f}\int \mcN(z;0,v_f)\mcN(z;\mu_z,v_z) \dd z
\end{align}
in which
\begin{align}
v_f = -\frac{f(0)}{f''(0)}.
\end{align}
Finally, recalling our assumption about the vector of true skills $\btheta^*$ (see Section~\ref{Sec:assumptions}) which implies that $\mu_z = \eta$ and $v_z=2v$, \eqref{prod.Gaussian} becomes
\begin{align*}
\Ex[f(z)] &\approx f(0)\sqrt{2\pi v_f} \mcN(0;\mu_z,v_z + v_f) \\
&= f(0) \sqrt{\frac{v_f}{2v + v_f}} \exp{\left[\frac{-\eta^2}{2 (2v + v_f)}\right]}. \numberthis
\end{align*}
Note that some pre-calculated values of $v_f$ and $f(0)$ used here to calculate \eqref{mean-h_k}, \eqref{mean-h_k^2}, and \eqref{eq:l_min} are presented in Table \ref{tab:appendix.a} for ease of use.

\begin{table}[!ht]
\centering
\caption{Identities used in the derivations for different functions $f(z)$.}
\label{tab:appendix.a}
\medskip
\begin{tabular}{@{}ccc@{}}
\toprule
$f(z)$ & $v_f$ & $f(0)$\\ 
\midrule
$\mcL(z)$                                          & $2$         & $1/4$      \\
$\mcL^2(z)$                                        & $1$         & $1/16$     \\
$-\sigma(z)\ln[\sigma(z)] - \sigma(-z)\ln[\sigma(-z)]$ & $4\ln(2)$ & $\ln(2)$\\ \bottomrule
\end{tabular}
\end{table}

\section{\normalfont\it Determination of \eqref{eq:optimum_step}}\label{Appendix.Step}

To derive an approximate solution for \eqref{eq:optimum-beta}, we resort to the Taylor series to expand \eqref{eq:v_k-final} around $\beta=0$, so that
\begin{align} \label{dk-taylor}
\ov{d}_k &\approx \ov{d}_k\Big|_{\beta=0} + \beta \left( \frac{\partial \ov{d}_k}{\partial \beta} \Big|_{\beta=0} \right) + \frac{1}{2} \beta^2 \left( \frac{\partial^2 \ov{d}_k}{\partial \beta^2} \Big|_{\beta=0} \right) 
\end{align}
in which
\begin{align} \label{dk-diff1}
\frac{\partial \ov{d}_k}{\partial \beta} \Big|_{\beta=0} &= -\frac{4 k \ov{h} M v}{M-1}
\end{align}
and
\begin{align} \label{dk-diff2}
    \frac{\partial^2 \ov{d}_k}{\partial \beta^2} \Big|_{\beta=0} = \frac{8 k \ov{h} M v}{M-1} \left[ \frac{2 \ov{h} (k - 1)}{M - 1} + \frac{\ov{h^2}}{\ov{h}} + \frac{1-1/M}{2 v} \right].
\end{align}
So, substituting \eqref{dk-diff1} and \eqref{dk-diff2} into \eqref{dk-taylor},  differentiating the resulting expression with respect to $\beta$, equating it to zero, and solving for $\beta$, one gets
\begin{align} \label{beta-taylor}
\beta_{\tr{o},k} &\approx \frac{1}{2} \left[ \frac{(1 - 1/M)}{2v} + \frac{\ov{h^2}}{\ov{h}} + \frac{2 \ov{h} (k - 1)}{M - 1} \right]^{-1}.
\end{align}
Next, in order to evaluate the accuracy of the approximate solution obtained, Fig.~\ref{fig:optimum_taylor} plots the results obtained from \eqref{beta-taylor} against those coming from the numerical solution of \eqref{eq:optimum-beta} as a function of $k$, considering different values of $v$ with $M=15$ and $\eta=0$. By doing so, we see that \eqref{beta-taylor} exhibits an important mismatch compared with the numerical results. Such a mismatch is more noticeable as $k$ increases, which leads us to infer that the contribution of the $k$-dependent term
\begin{align}\label{taylor-last-term}
    \frac{2 \ov{h} (k - 1)}{M - 1}
\end{align}
is not properly balanced. Moreover, for the special case of $k=1$, an analytical solution for \eqref{eq:optimum-beta} can be derived exactly as 
\begin{align} \label{beta-k=1}
\beta_{\tr{o},1} &= \frac{1}{2} \left[ \frac{(1 - 1/M)}{2v} + \frac{\ov{h^2}}{\ov{h}} \right]^{-1}
\end{align}
in which \eqref{taylor-last-term} does not appear (as expected); thereby, it reinforces the idea that the contribution of the term dependent on $k$ in \eqref{beta-taylor} must be revised. 
After an extensive analysis of different scenarios, we observed that $\ov{h}$ in \eqref{taylor-last-term} is too large and can be replaced by $\ov{h^2}$; as a result, a more accurate solution than \eqref{beta-taylor} to calculate the optimal step-size value follows as \eqref{eq:optimum_step}.

\renewcommand{\thefigure}{\arabic{figure}}

\begin{figure}[t]
\centering
\includegraphics[scale=1]{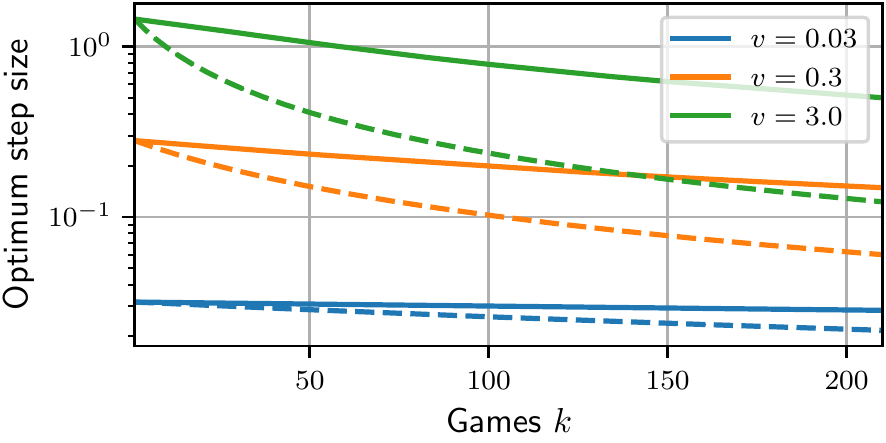}
\caption{Optimal step-size value $\beta_{{\tr{o}}, k}$ obtained from the numerical solution of \eqref{eq:optimum-beta} (solid lines) and the approximate solution \eqref{beta-taylor} (dashed lines) as a function of the number of games $k$, considering different values of $v$ with $M = 15$ and $\eta = 0$. Curves for other values of $\eta$ are omitted since this variable does not affect \eqref{beta-taylor} significantly.}
\label{fig:optimum_taylor}
\end{figure}

\bibliographystyle{IEEEtran}
\bibliography{references.bib}

\end{document}